# High-Performance Vision-Based Tactile Sensing Enhanced by Microstructures and Lightweight CNN

Mayue Shi[1]*, Yongqi Zhang[1], Xiaotong Guo[1] and Eric M. Yeatman[1]

[1]Department of Electrical and Electronic Engineering, Imperial College London, London, UK.

*Email: m.shi16@imperial.ac.uk.

## Abstract:

Tactile sensing is critical in advanced interactive systems by emulating the human sense of touch to detect stimuli. Vision-based tactile sensors are promising for providing multimodal capabilities and high robustness, yet existing technologies still have limitations in sensitivity, spatial resolution, and high computational demands of deep learning-based image processing. This paper presents a comprehensive approach combining a novel microstructure-based sensor design and efficient image processing, demonstrating that carefully engineered microstructures can significantly enhance performance while reducing computational load. Without traditional tracking markers, our sensor incorporates an surface with micromachined trenches, as an example of microstructures, which modulate light transmission and amplify the response to applied force. The amplified image features can be extracted by a ultra lightweight convolutional neural network to accurately inferring contact location, displacement, and applied force with high precision. Through theoretical analysis, we demonstrated that the micro trenches significantly amplified the visual effects of shape distortion. Using only a commercial webcam, the sensor system effectively detected forces below 5 mN, and achieved a millimetre-level single-point spatial resolution. Using a model with only one convolutional layer, a mean absolute error below 0.05 mm was achieved. Its soft sensor body allows seamless integration with soft robots, while its immunity to electrical crosstalk and interference guarantees reliability in complex human-machine environments.

# 1. Introduction

Soft tactile sensors is a key aspect of humanoid robots and other intelligent interactive systems [1]. It mimics the human sense of touch, enabling the detection of stimuli such as pressure, vibration and temperature [2]. Many applications have been developed in various fields [3], such as robotics [4], healthcare monitoring [5], [6], and prosthetics [7], [8]. However, current soft tactile sensing technologies encounter significant challenges, especially in sensitivity, resolution and durability. Additionally, for large-scale sensing, which is critical for robots and prosthetics applications, current sophisticated sensing systems based on sensor arrays often have to face dramatical increase of complexity, electrical crosstalk and interference, and usually unacceptable cost. It is necessary to develop innovative solutions to enhance tactile sensing capabilities.

Vision-based tactile sensors (VBTSs), which utilise visual information to infer multimodal tactile properties, are promising to address the limitations in current soft tactile sensing. They can effectively detect location and magnitude of pressure, characterise surface textures, and operate in a wide range of scenarios. This innovative approach has been investigated for robotic perception [1], offering rich information, high robustness and adaptability, low cost, and multimodal sensing capability [9]. Moreover, VBTSs are compatible with many computer vision methods, such as convolutional neural networks [2]. Recently, the rapid development of artificial intelligence technology has effectively boosted the development of VBTSs [10].

The structure of a VBTS often contains a sensor body and camera, among which the design of sensor body is critical for the improvement of perception performance, while the camera module remains relatively standardised. The optimisation design of the sensor body generally involves several aspects, including novel structures, advanced materials and fabrication processes. With the rapid improvement of hardware designs, the last decade has witnessed the development of the VBTS towards miniaturization, high performance and multimodality [11]. Current VBTSs are mainly based on two principles: displacement of markers with deformation of the sensor [12], [13], [14], [15], [16], [17], and irregular refraction and reflection of light induced by surface deformation [18], [19], [20]. For example, GelForce is a well-known marker-based sensor designed to measure the surface traction field on a robotic hand [15], [16]. It has two layers of spherical markers embedded within a transparent silicone rubber sensor body. In addition, the TacTip family realises superresolution tactile sensing for localization tasks with a 3D printed sensor body with markers, utilising pins on the sensor body to mimic the function of intermediate ridges within the human fingertip [13]. In contrast, early GelSight sensors captured high-resolution geometry without the uses of markers. They utilised photometric stereo to reconstruct the depth map and to infer local force [20], [21]. Later versions of GelSight introduced markers in the reflective membrane to enhance the capability to measure force and torque loads [19], [22]. In addition to advances in sensor body design, improvements in camera systems and advanced image processing methods further improved the performance of VBTSs. A multi-camera design was presented by Trueeb et. al [23] to overcome the limited field of view inherent in single-camera systems. By monitoring the particle patterns in the sensor body, contact force distribution was measured with a machine learning method. A transfer learning algorithm was also explored to reconstruct normal force distribution and facilitate knowledge transfer across different sensor gels. [24]. Evetac is a novel event-based optical tactile sensor designed to enhance temporal resolution and detect vibrations up to 498 Hz, leading to more adaptive and precise robotic manipulation. It processed measurements online at 1000 Hz, using efficient algorithms to track deformation and reduce data rates compared to traditional RGB sensors [25]. Recently, large-scale tactile sensing methods for robotic manipulators have gained attention. For instance, a study demonstrated the use of electroluminescent panels combined with a deformable soft skin to localize forces and estimate their magnitude in multipoint contact scenarios, achieving millimetre-level localisation accuracy [26].

Despite some studies exploring innovative structural designs for VBTSs, the potential performance benefits of integrating microfabrication approaches into these sensors have not been investigated. These approaches for both soft and rigid materials, however, are well-established and proven effective in enhancing sensor performance in various micro-electromechanical systems (MEMS). Recently, we have successfully developed the fabrication process of a microstructure-enhanced VBTS [27], and initially demonstrated its potential to achieve high precision sensing. In the current study, we comprehensively developed this sensor system, named MicroSight, from hardware to data processing to infer the displacement, location and magnitude of applied force with an ultra-light CNN model (**Figure 1**). The performance of the MicroSight sensor was carefully evaluated. With micromachined cross-shaped structures on the stretchable sensor body and the computer vison method developed in this paper, the properties of the applied force can be measured with high precision across a detection range as low as the millinewton level.

# 2. Results

### 2.1. Structural design and working principle

As shown in **Figure 1**a, the stretchable sensor body was composed of a transparent polydimethylsiloxane (PDMS) layer and a thin black composite layer. A grid of 5×5 cross-shaped trench structures was precisely fabricated using ultraviolet laser cutting, with a depth of around halfway through the sensor body film (**Figure 1**a and **Figure 2**c). The distance between adjacent cross centres is 2 mm, with each cross measuring 1.5 mm in both length and width. The sensor body is clamped using a 3D-printed frame with a 1.6x1.6 cm square window. The data processing pipeline of the MicroSight system, as depicted in **Figure 1**c, involves image acquisition and preprocessing, followed by lightweight CNN processing to learn the patterns and features necessary for prediction of displacement in depth (*z*-coordinate), applied force and location coordinates (*x*, *y*). As shown in the figure, the comparison between true values and CNN predictions demonstrates the system's high accuracy.

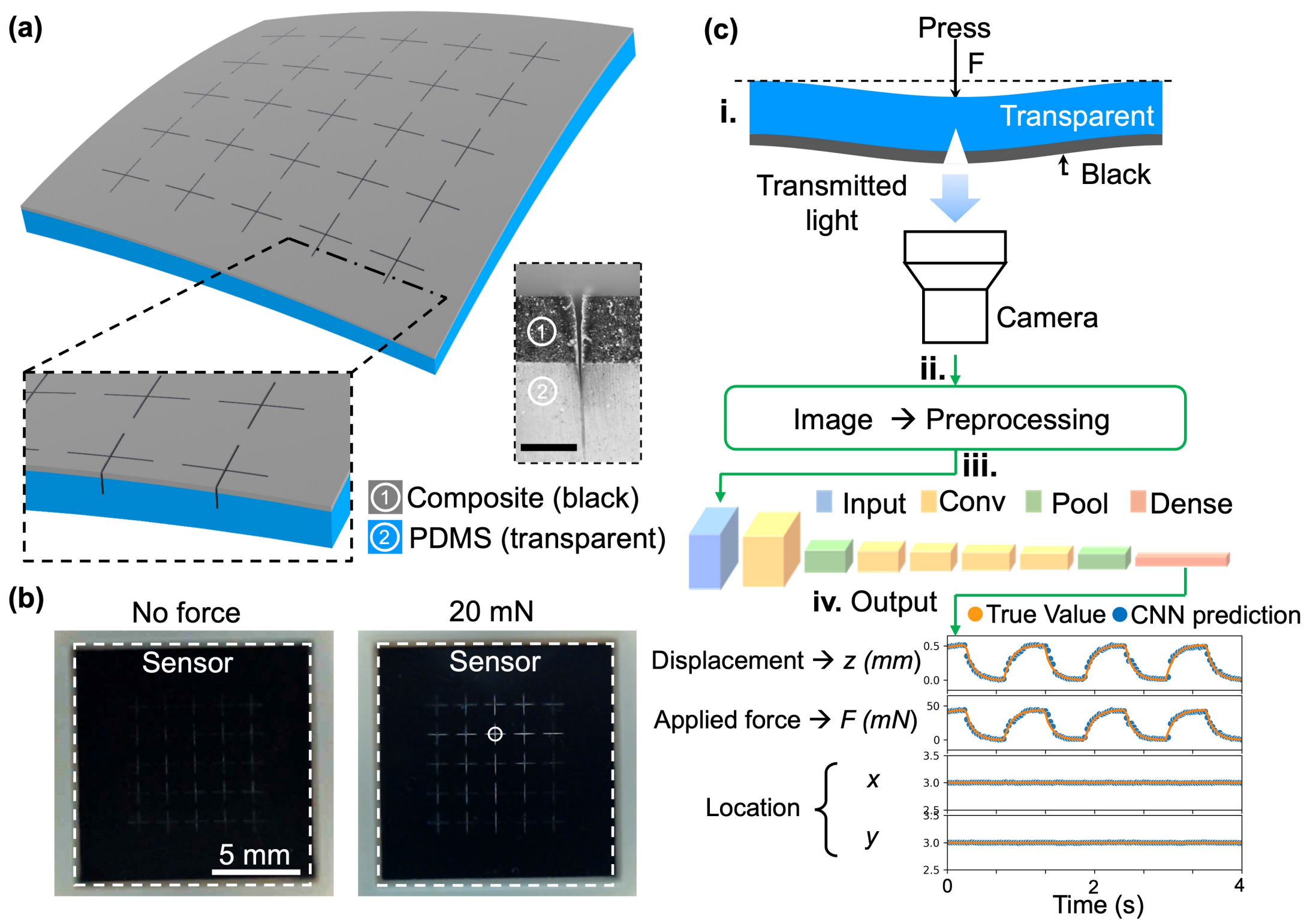

**Figure 1**: Structural schematic diagram and system design. (a) Structural design of the sensor body, and a sectional view showing trench structures in the film (scale bar: 200 μm). (b) Optical images of the sensor under different applied forces. Left: no force is applied. Right: a 20 mN force is applied at the circled cross centre in row 2, column 3. Diameter of contact area: 1 mm. (c) System design. i. Sensor body with micro trench patterns modulates light under pressure. ii. Camera captures images with transmitted light patterns and then iii. the images are pre-processed, and iv. feed into a light-weight CNN for analysis. v. Four outputs of CNN represent vertical displacement, applied force, and the location of the contact point on the sensor surface.

**Figure 2** illustrates the working principle of this MicroSight sensor that leverages the deformation of microstructures to measure applied force. The detailed structure of a micro trench is characterised in **Figure 2**c. The sensor body was fabricated with two materials:

transparent PDMS elastomer and black graphite/PDMS composite. When applying pressure on the transparent side, the deformation of the film leads to an increase of micro trench width and luminous flux, and amplified by the high dept-to-width ratio of microtrench structure. Then, images are captured by a camera under the sensor body. For instance, the images of our sensor body clearly showed the distinguishable features under an applied force of only 20 mN (**Figure 1**b), which indicates its high sensitivity. With the diaphragm structure and elastic sensor body, the sensor recovers spontaneously after deformation.

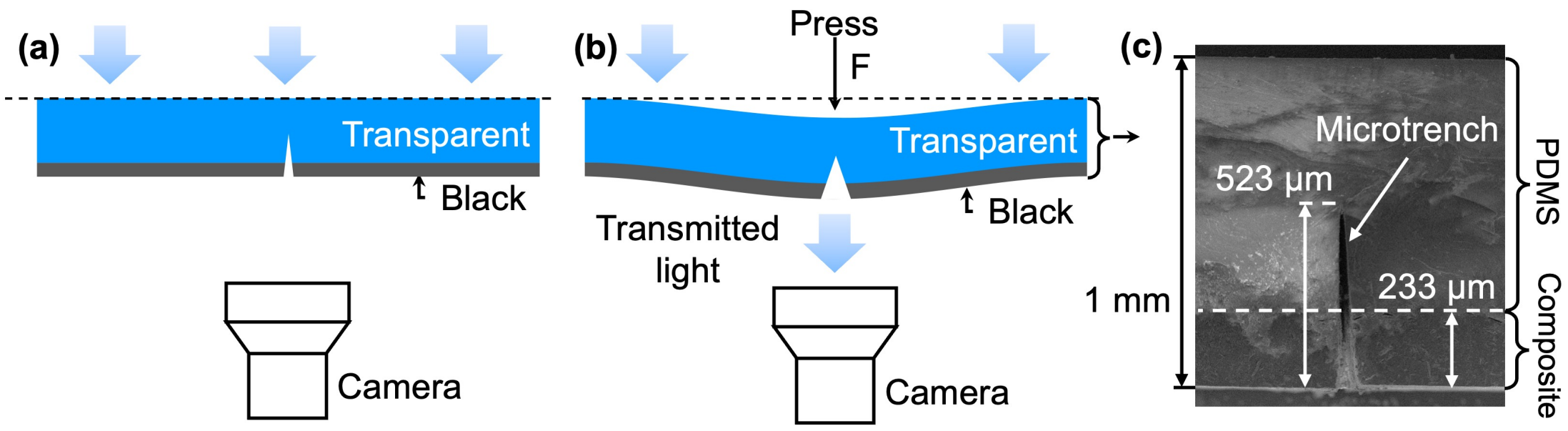

**Figure 2**: Principle of structure-enhanced vision-based sensor. (a) No applied force; (b): with 20 mN applied force. Deformation of surface under pressure allows light to transmit from open micro trenches, forming specific patterns captured by the camera. (b) Characterisation of a micro trench with SEM microscope.

### 2.2. Theoretical modelling and simulation

We developed a comprehensive mathematical model to analyse the bending behaviour of a thin PDMS film with fixed boundaries and uniformly distributed V-shaped notches (**Figure 3**). With this model, we derived the extension of trench width ($\Delta d$) under an applied force, which is critical for modulating light transmission and amplifying the variation. This model simplifies the complex behaviour of the sensor body film by representing it as an array of fixed-fixed beams. Each beam bends independently in response to an applied load, assuming that cross-sections remain undeformed during bending. A detailed modelling process is presented in the **Supplementary Note 1**.

As a result, for location $i$ ($i \in \{A, B, C, D, E\}$), the deformations of the notch $\Delta d_i$ (**Figure 3**b) can be expressed as:

$$\Delta d_i = \frac{3M_i}{d_0 E\alpha} \tag{3}$$

Here, $M_i$ is the bending moment at location $i$; $d_0$ denotes the original trench width without applied force; $E$ denotes the effective Young's Modulus of the film, determined through experimental calibration; $\alpha$ is the geometric depth of the trench, also approximately half of the entire film thickness ($h \approx 2\alpha$) in our sensor.

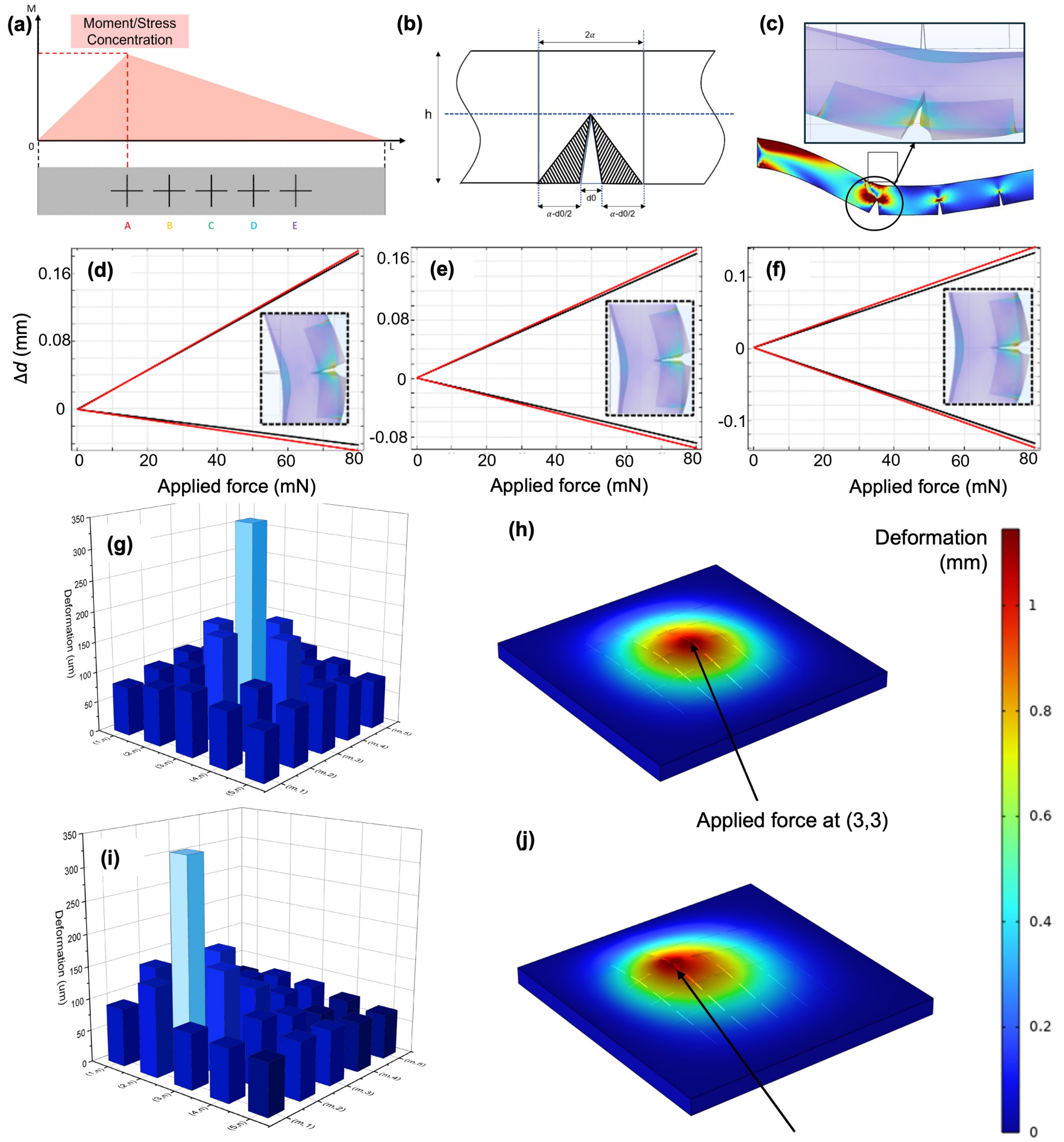

**Figure 3**. Modelling and simulation analysis. (a) and (b) modelling of the sensor body structure. (a) provides the illustration of stress concentration. (c) Stress distribution at location A under applied force based on FEA analysis. (d)-(f) present the deformation pattern of the notched beam at locations (d) A, (e) B, and (f) C simulated using COMSOL Multiphysics. The load range analysed was from 0 to 80 mN. Comparison between the simulated (black) and modelled (red) deformations demonstrates our model consistently predicted a slightly larger deformation at both edges, with discrepancies within 13% across varying applied loads. (g)-(j) The average diagonal length (g and i) and deformation (h and j) of all 25 micro trenches under an applied force of 60 mN at location (3, 3) and location (2, 2).

To validate the mathematical model, we compared the FEA results with the analytical predictions (**Figure 3**e,f,g). Using COMSOL Multiphysics, we performed simulations over a range of applied loads (0 to 80 mN), observing deformation patterns that closely matched our theoretical predictions. Further FEA was conducted (**Figure S2**) to model the deformation and stress distribution of the sensor body under a 60 mN force applied at the central location. The analysis revealed significant deformation around the contact point, affecting the micro trench's luminous flux. To investigate the change of microtrench width across the 5×5 array, simulations were conducted for two scenarios (**Figure 3**g to j): a 60 mN force applied at the central location (3, 3) and at an off-centre location (2, 2). Similar to our previous study [27], we use the average diagonal length of the quadrilateral at the cross center to represent the deformation of the microtrench under an applied force. **Figure 3**g and i present the average diagonal length at the 25 cross centres for two scenarios. Both results indicated significantly larger variation at the contact point compared to other cross centres, which results in amplified light transmission and enhanced visual features benefiting further feature extraction.

Furthermore, the deformation results reveal that deformation patterns were radially symmetrical across the 5×5 array when the central location (3,3) is pressed (**Figure 3**h). Locations closer to the point of contact experience greater deformation compared to other cross centres, but with less contrast compared to the average diagonal length results, shown in **Figure 3**g. In **Figure 3**j, where the off-center location (2,2) is pressed, points such as (5,1), (4,4), and (5,5), which are farther from the contact point, experience less deformation compared to their counterparts when the central location (3,3) is pressed.

### 2.3. Experimental set-up and data collection

The fabrication of the sensor body involves two key steps: the fabrication of a multi-layer elastomer film followed by a high-precision patterning process using ultraviolet (UV) laser cutting. During the multi-layer film fabrication, spin coating was primarily used to ensure uniformity and precise control over the film thickness. For the laser cutting process, a high-precision UV laser cutter was employed, achieving a high aspect ratio microtrench structure with an opening width of approximately 40 μm. Detailed fabrication procedures of the sensor body are provided in **Figure S4** and **Methods** [27]. According to SEM measurement (**Figure 1**e), the total thickness of the sensor body is 1mm, while the composite layer is ~230 μm. The depth of the micro trench was ~523 μm.

The performance of the MicroSight sensor was tested with a bespoke experimental setup, as shown in **Figure S5** and **Methods**. A commercial Logitech C922 camera (1920×1080 pixels resolution) was used to capture images at 30 frames per second (FPS) during force applications. We tested the sensor with a single contact point located at the centre of all crosses. Displacements of 0–0.5 mm, 0–1.0 mm, and 0–1.5 mm were applied against the sensor surface, with all 25 sensing points tested repeatedly under consistent characterization conditions and ambient lighting. The displacement of linear motor was synchronously recorded. It is obvious that images captured under these conditions showed clear distinctions. The results from three typical testing points of (3,3), (1,4) and (4,4) are shown in **Figure 4**b to d. The images indicate that crosses close to the contact point had high luminous flux and wide trenches, which is consistent with the modelling result. Additionally, the deformation of the film led to the displacement of crosses in images. This result was predicted by finite element simulation as well.

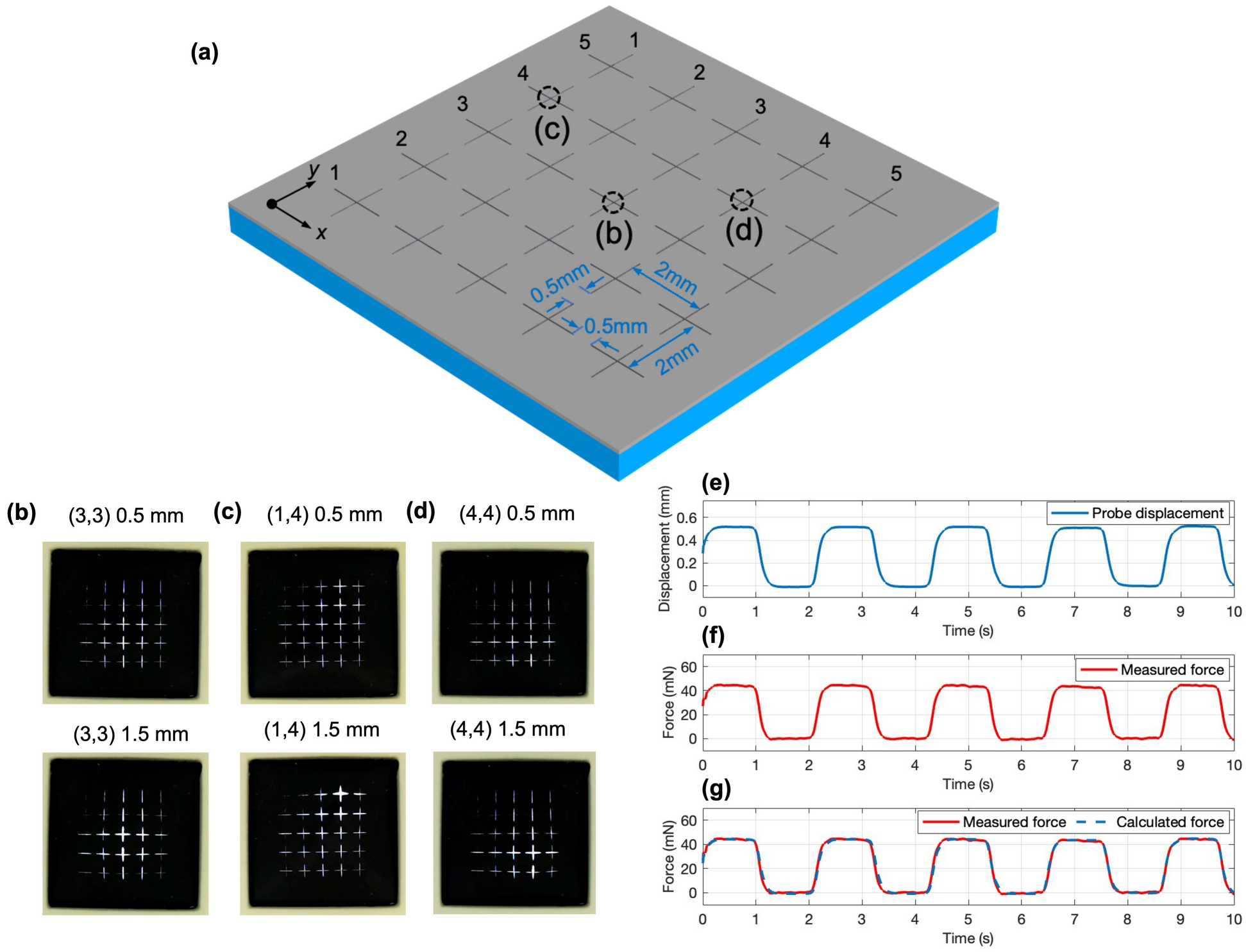

**Figure 4**: Sensor response to applied force and force calibration. (a) Structure and parameters of the sensor body. (b), (c) and (d) Optical images captured using camera at three typical locations. The data above each image indicates the contact location and the displacement of the linear motor toward the sensor. For example, (3, 3) indicate the applied force was at the cross centre in row 3, column 3. Diameter of contact area: 1 mm. (e), (f) and (g) represent the force - displacement calibration of a testing point (3,3). (e) Displacement and (f) force was measured to calibrate the coefficient $k_{33}$. (g) Comparison of the measured force and the calculated force according to the displacement over time ($k_{33}$=85.4 mN/mm).

A simple quantitative image analysis is shown in **Figure S3**. We captured images under three displacement conditions at the testing point of (3,3), and calculated the average brightness of the square central region (60×60 pixels) to reflect the response of the sensor to different vertical displacements. The calculation is performed using value channels of images in the HSV (hue, saturation, value) colorspace, in which the value describes the brightness. The figure clearly indicated that the average brightness increased with the displacement. Specifically, when the displacement varied from 0.5 mm to 1.5 mm, the average brightness increased to 3.2 times that of 0.5 mm displacement. It suggests the outstanding sensitivity of our sensor, even when evaluated solely through the metric of average brightness. If we only focus on the cross-trench region, the variation of brightness was even higher. This is especially beneficial for further in-depth image processing based on machine learning models to achieve high precision.

The applied force can be calculated based on vertical displacement ($z$-direction) based on a simple linear relationship. According to theoretical analysis and simulation [27], the relationship can be approximately expressed as $F(z) = k_{xy} \times z$, where the coefficient $k_{xy}$ represents the proportionality between vertical displacement ($z$) and the applied force ($F$) at location ($x$,$y$). To calibrate the coefficient $k_{xy}$ , we measured the applied force with a high-precision scale, as described in **Methods**. As an example, the displacement results from the point (3,3) is shown in **Figure 4**e and f. According to the measured force and probe displacement, a linear curve fitting was performed, and the resulting $k_{33}$ value was calibrated as 85.4 mN/mm. The coefficient of determination $R^2$ was calculated as 0.99 at this point, which validated the linear model is a good approximation of force-displacement relationship, as shown in **Figure 4**g.

### 2.4. CNN model architecture and evaluation

We have developed an ultra-light CNN model architecture to infer the location and magnitude of applied force, as shown in **Figure 5**. The optical information collected by the camera was pre-processed before being passed to the CNN model (**Figure 1**c). Specifically, the collected dataset included 225,000 frame images, consisting of 25 testing points, each subjected to 3 sequential displacement ranges starting from 0. For each displacement range, 3000 images were recorded, corresponding to 100 seconds at 30 FPS of data collection. After this, each image was processed by retaining only the value channel, and then cropped into 25 small images of 60×60 pixels, centred at the cross points. These cropped images were then input into CNN as 25 separate channels (**Figure 5**a). To further enhance the diversity of the training data, data augmentation techniques were employed, including random translations within 1% of the image scale and rotations up to 1.8°. This approach helped to mitigate overfitting and improve the model's robustness and generalisation capabilities. The training process was conducted on a high-performance computing cluster, allowing for efficient handling of the large dataset. Before splitting the dataset, it was shuffled to enhance randomness and prevent order bias. It was then divided into 70% for training, 15% for validation, and the remaining 15% for testing.

We tested models comprising only 1, 3, 5 and 7 convolutional layers. For instance, the model with 5 convolutional layers (**Figure 5**b, denoted as CNN_5) has a total of 406k parameters. The image processing pipeline begins with preprocessing layers that rescale the pixel values. The image matrix values were rescaled to a range of 0 to 1 by dividing by 255, standardising the inputs and facilitating faster convergence. The convolutional layers progressively extract features, each followed by batch normalisation and ReLU activation to enhance learning efficiency and stability. The first convolutional layer outputs a shape of (54, 54, 64), while subsequent layers further refine the features, culminating in a (10, 10, 128) output after the fifth layer. The architecture also includes max pooling and average pooling layers to reduce spatial dimensions and computational complexity. Finally, the output is flattened and passed through a dense layer to predict the x, y, and z coordinates of the applied force.

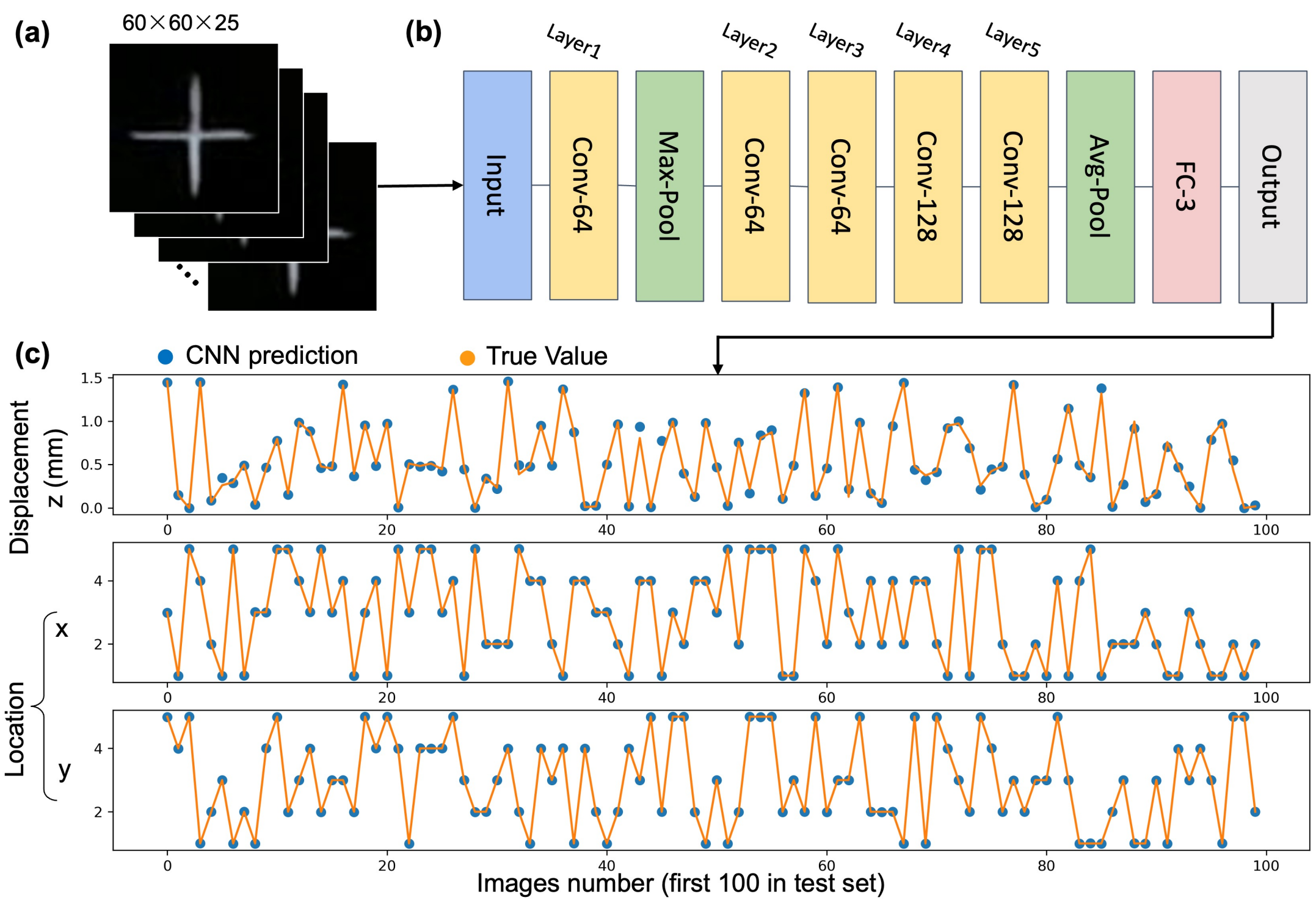

**Figure 5**: Architecture of the CNN model. (a) pre-processed image as input. (b) Model architecture with 5 CNN layers (CNN_5). (c) A comparison between the predicted values by the CNN model and the true values for the first 100 images in test set.

**Figure 5**c demonstrates the prediction results of 100 sample images in the test set with our CNN_5 model, which reflects that the simple lightweight model performed well in predicting the applied forces' $x$, $y$, and $z$-coordinates. The predictions for the $x$ and $y$-coordinates are particularly accurate. The displacement predictions are slightly less precise but still follow true values closely for most samples. These results indicates that the CNN model is effective for predicting the location and displacement. As demonstrated previously, the magnitude of applied force can be derived from the displacement using a simple linear relationship.

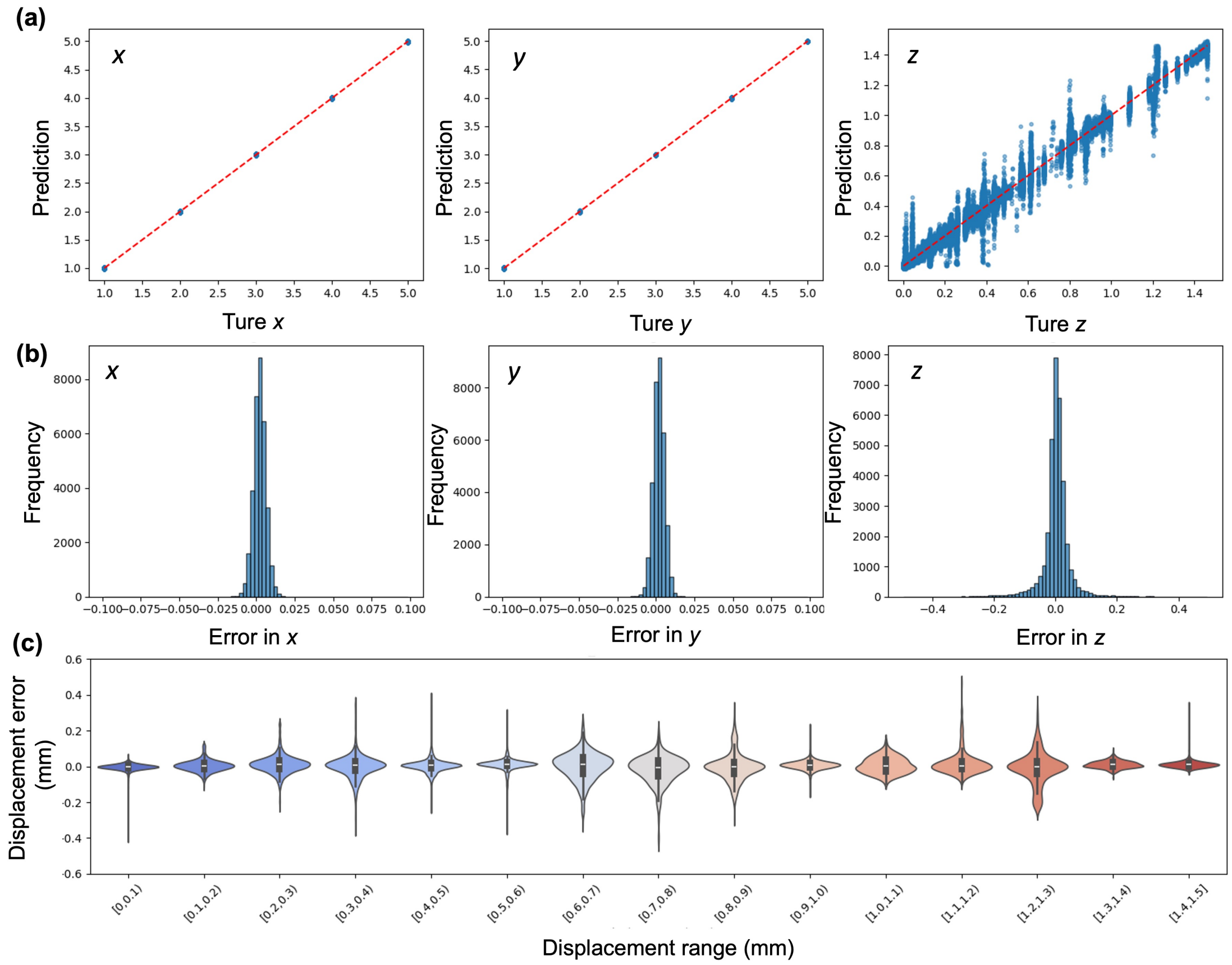

**Figure 6**: Evaluation of model performance. (a) Scatter plots comparing the predicted values to the true values. (b) Histograms of the prediction errors. (c) Distribution of displacement prediction errors across different ranges of actual values.

To systematically evaluate the model's performance, we conducted a series of tests and statistic analysis using the test set, containing 15% of the images of the entire dataset. The results are shown in **Figure 6** and **Figure 7**. In **Figure 6**a, each subplot shows a clear correlation between the predicted and true values, indicated by the close alignment of the points along the red dashed line, which represents that predicted values equal to true values. Consistent with the initial analysis of the results in **Figure 5**c, the scatter plot for the *x*- and *y*- coordinates (left subplot) demonstrates an outstanding performance with most points falling on the line, indicating high accuracy in predicting the location of the applied force. For the *z*-direction or displacement (right subplot), there is more spread around the line, indicating higher variance in the predictions. However, the overall trend still shows a strong correlation between the predicted and true values.

Histograms of the errors ($r_i$, where $i$ represents the sample index) are shown in **Figure 6**b. These errors, defined as as $r_i = y_i - \hat{y}_i$, reflect the differences between the sensor outputs ($\hat{y}_i$) and true values ($y_i$). The residuals for the $x$- and $y$-coordinates (left and middle subplots) are narrowly distributed around zero, with the majority of residuals being very close to zero, confirming the model's high accuracy in location prediction. The residuals for the $z$-coordinate (right subplot) have a wider distribution, indicating lower prediction accuracy. Nevertheless, 95% of the residuals fall within a narrow range, and the distribution peak is still centred around zero, suggesting that the predictions are unbiased.

To evaluate the sensor's ability to measure displacement within different measurement ranges, we visualised the distribution of displacement prediction errors using a violin plot (**Figure 6**c). This violin plot visualises the error distribution across different bins of actual values from 0 to 1.5 mm. Overall, these bins have a median error (white dot) near 0, indicating no significant systematic bias in predictions. Prediction errors concentrated around 0 in most bins, meaning that predictions align closely with actual values. The MicroSight sensor maintains high accuracy even for small displacements, with no evident performance degradation. Interestingly, for the smallest and largest bins, the direction of larger errors is notably asymmetric and occurs in opposite directions for these two bins. This phenomenon is understandable, as the model inherently learns to predict values within the range of sample data, which is within the 0–1.5 mm, rather than exceeding these limits. This also indirectly suggests that the model has effectively learned the characteristics of the sample data.

Furthermore, we analysed the *x*-*y* error distribution to assess location precision, as shown in the **Figure 7**. This figure illustrates the *x*-*y* error patterns at 25 positions. In each subplot, the blue dots indicated individual error measurements, and the black error ellipses represented the 95% confidence interval. This figure shows that most distributions concentrated around the origin, indicating relatively low errors. However, the spread of points in certain positions shows greater variability in error, suggesting slight localised differences in tactile sensing accuracy across the sensor.

In general, this model containing 5 CNN layers demonstrates excellent performance in predicting the location and displacement, particularly in location prediction. The slightly higher residual in the *z*-coordinate predictions suggests that there may be room for further optimisation of the model or the training process to improve accuracy in this dimension. Despite this, the CNN_5 model's performance is robust and effective for precise tactile sensing.

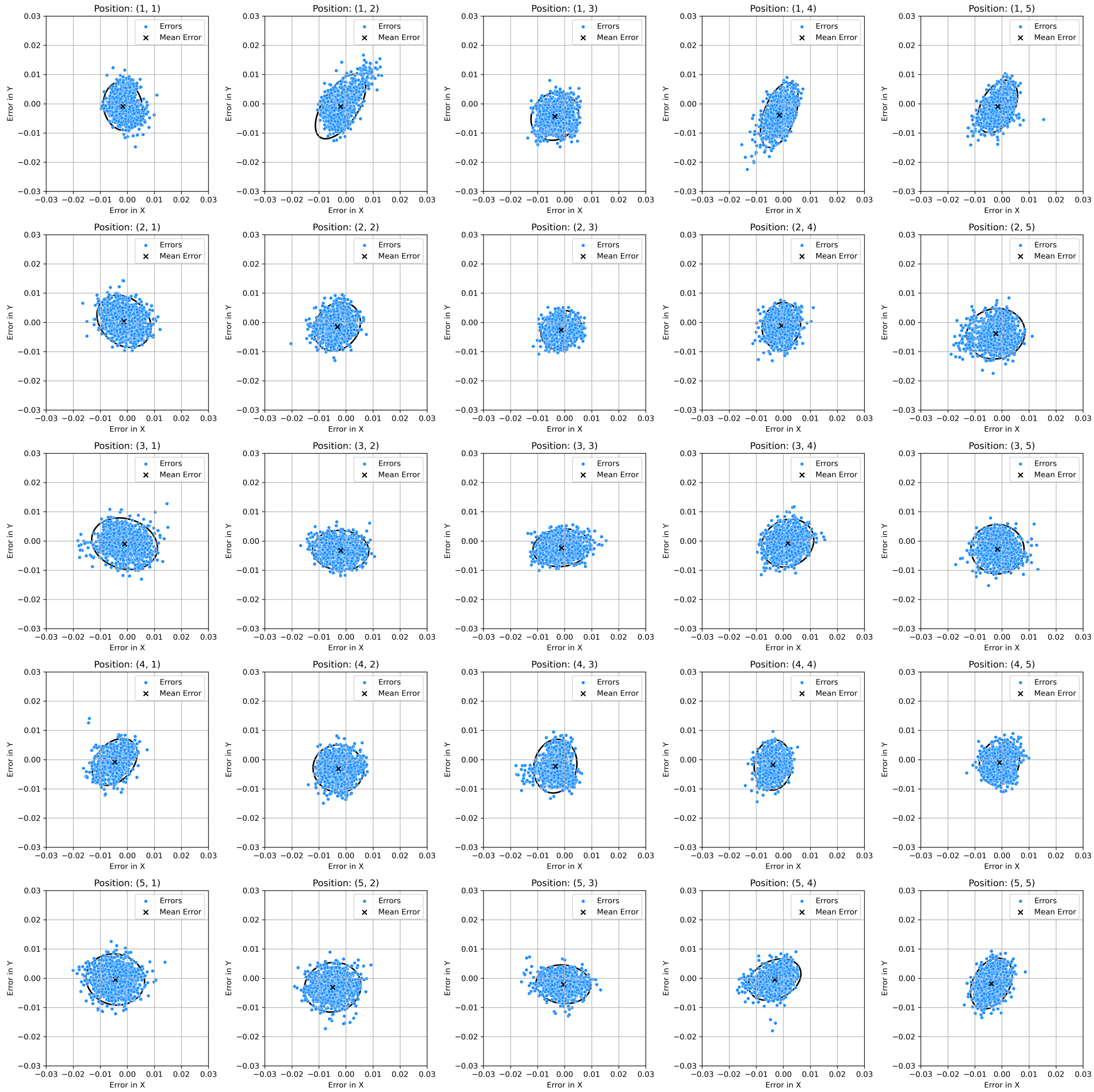

**Figure 7**: Error distribution for location resolution across different positions of the sensor.

To evaluate the impact of CNN depth on sensor performance, we systematically evaluated lightweight models with different numbers of convolutional layers, ranging from 1 to 7. **Figure 8**a to c presents the mean squared error (MSE) and $R^2$ values for different CNN models with varying numbers of layers (1, 3, 5, 7 layers). **Table S1** shows a complete comparison of performance metrics for different models. The results indicates that increasing the number of CNN layers generally improves the model's performance, as indicated by decreasing MSE and increasing $R^2$ values. While performance improves up to 5 layers for the x coordinate and continues to improve up to 7 layers for the y and z coordinates, there may be diminishing returns, particularly for the x and y coordinate. However, even the CNN_1 model with only 134k parameters achieved a relatively low MSE and an $R^2$ above 0.98. The model file saved in the Hierarchical Data Format (HDF5) is only 1.6MB, making it suited for deployment on memory- and power-constrained edge computing devices.

Overall, our model achieved high precision. Due to the introduction of microstructures, the light flux variations caused by surface deformations on the sensor were amplified effectively. This amplification makes changes in graphical features more pronounced, allowing even shallow neural networks to effectively capture these features. Consequently, this reduces the computational and storage resource requirements, enabling efficient feature extraction with lightweight CNN architectures. As a result, it has the potential for deployment on edge devices providing high-precision sensing while also reducing the computational burden on robots. This dual advantage highlights the efficacy of our design in balancing performance with computational efficiency. **Table S2** provides a comparative analysis of our results against other VBTS studies. The table highlights differences in optical systems, image resolutions, processing models, and errors in location and force/depth measurements.

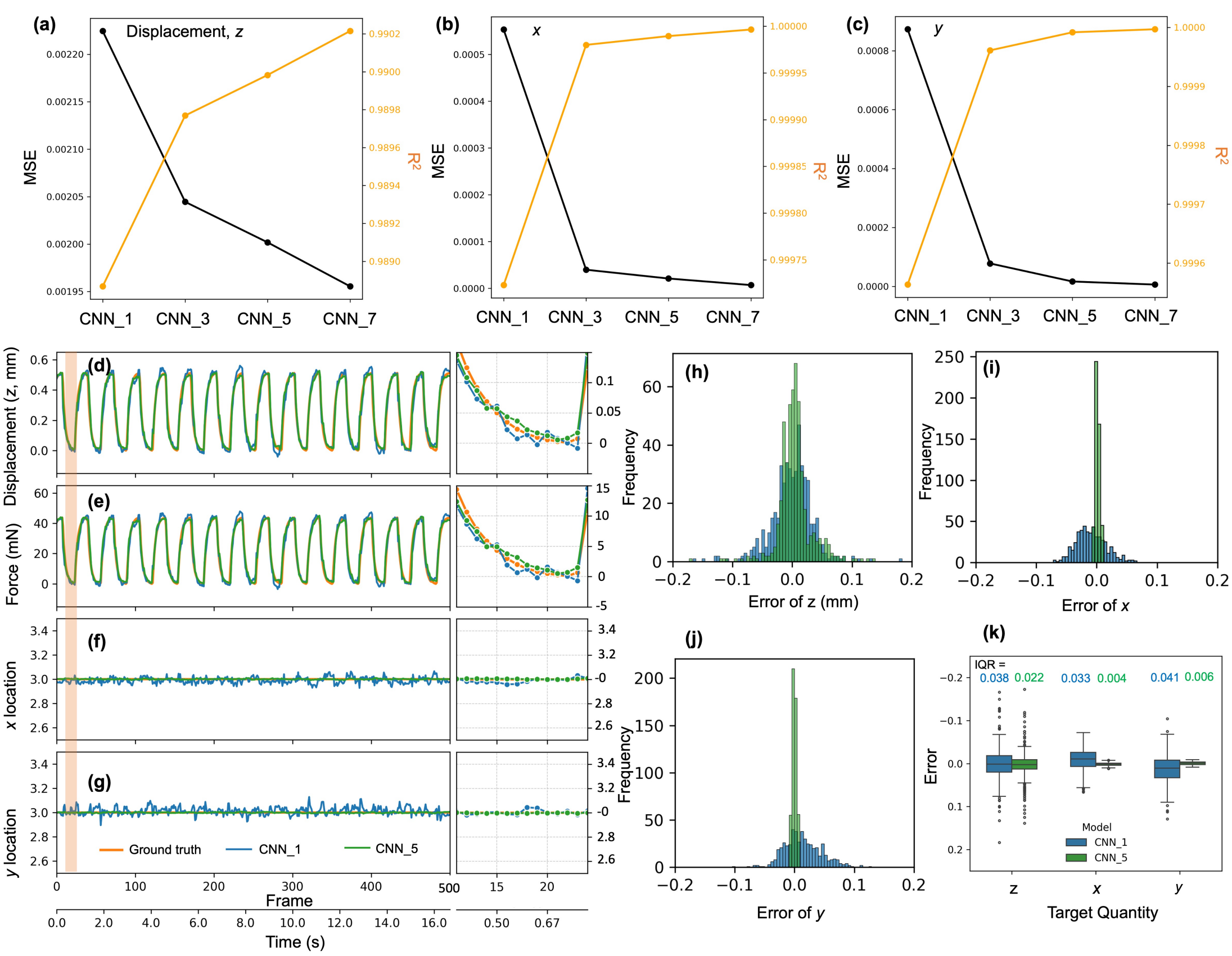

**Figure 8**: Comparison of performance for different CNN models. (a) – (c) Performance metrics for models with varying numbers of convolutional layers. (d) – (k) present the experimental testing results and error analysis at location (3,3) using CNN_1 (blue) and CNN_5 (green). Specifically, ground truth and model predictions for (d) displacement and (e) calculated force in *z*-direction, (f) *x* and (g) *y*-coordinate results are shown. The shaded amber region highlights a low-force segment, with zoomed-in views provided on the right. (h) – (k) present the error distributions and boxplots to evaluate prediction accuracy for this sensor system.

To assess the sensor system's capability in continuous multimodal measurements at a specific location, we conducted tests at point (3,3) using models of varying depths (CNN_1 and CNN_5). **Figure 8**d to k show the results of 500 consecutive frames captured at a rate of 30 frames per second. The applied force was calculated with $k_{33}$=85.4 mN/mm, as calibrated previously. The predicted displacement and force from both CNN_1 and CNN_5 aligned closely with the ground truth. This strong correlation validated the system's high temporal accuracy and reliability during continuous testing. Both models achieved relatively accuracy, but CNN_5 yields visibly smoother and more stable predictions , as emphasized in the magnified zoomed-in views. Notably, the zoomed-in segment corresponding to the low-force region clearly demonstrated that the MicroSight system is capable of accurately capturing force, displacement, and positional changes even at low-force levels below 5 mN and displacements below 50 μm. These results were consistent with the comprehensive evaluation in **Figure 6**c and **Table S1**. Traditionally, VBTS systems have exhibited limited performance along the optical axis of the camera—i.e., in the direction perpendicular to the imaging plane (z-direction). In contrast, our sensor shows robust sensing performance in the z-axis as well. This improvement highlights the effectiveness of the integrated microstructure design in enhancing the overall performance of VBTSs.

The error histograms in panels (h)–(j) further quantify model accuracy. CNN_5 shows tighter and more symmetrical error distributions centered around zero for all axes, indicating improved consistency over CNN_1. Finally, **Figure 8**k shows the boxplot for sensing errors for displacement, x, and y locations. The result reflected that the model CNN_5 consistently performs better than CNN_1, as indicated by its lower interquartile ranges (IQRs).

# 3. Conclusions

In this paper, we propose a high-performance multimodal tactile sensing paradigm that integrates microstructures and computer vision. Based on this paradigm, we designed and implemented a MicroSight sensor using microfabricated microtrenches and lightweight CNNs, and systematically evaluated its performance. This innovative sensor design incorporates a black composite layer with 25 cross microtrenches, modulating the light luminous flux based on the applied force. Camera-captured images of the sensor body serve as the input for our ultra-light CNN model architecture to accurately infer the location, displacement and magnitude of applied force. We evaluated the models with 1 to 7 convolutional layers and found that the models effectively extract relevant features from the input images, facilitating precise predictions. The model architecture balances computational efficiency with high accuracy, making it suitable for real-time applications in IoT and edge-computing. Our MicroSight sensor demonstrated remarkable performance, providing high accuracy even within the millinewton-level force range while achieving millimeter-level spatial resolution. Its compatibility with computer vision methods enhances its potential for integration with and further enhanced by advanced AI systems. Furthermore, its soft sensor body makes it highly promising for seamless integration into advanced soft robots and wearable electronics.

In future research, this multimodal tactile sensing paradigm can be further developed and expanded for broader applications. These cross-shaped micro trenches in this work has demonstrated that well-engineered microstructures can effectively enhance tactile sensing. More complex three-dimensional geometries, achievable through MEMS fabrication, hold significant potential for further improving the performance. Additionally, the existing sensor design holds potential for multi-point tactile sensing by collecting and training on multi-point datasets. By integrating innovative sensor design, advanced microfabrication, and powerful machine learning, this study establishes a foundation for microstructure-enhanced vision-based tactile sensing. Our theoretical and experimental results underscore its potential to drive significant advancements in tactile sensing technologies, paving the way for next-generation applications in robotics and human-machine interfaces.

# 4. Methods

**Device fabrication**

We first fabricated a thin PDMS substrate film. The PDMS base and curing agent (Dow Corning, SYLGARD 184 silicone elastomer) were mixed in a ratio of 10:1 (w/w), after which the mixture was spin-coated on a plate before being transferred to an oven for curing at 70 °C for 2 hours. To prepare the graphite/PDMS composite, 10 wt% graphite (Sigma-Aldrich, 282863) was fully mixed with uncured PDMS (base: curing agent = 10:1, w/w) to form a black composite, enhancing image contrast. Toluene was added to the mixture to assist uniform dispersion and adjust its viscosity. The mixture was spin-coated on the transparent PDMS film (300 rpm, 60 s) to ensure uniformity. The film was transferred to an oven at 70 °C for 4 hours. After preparing double-layer elastomer film, the micro trenches were fabricated with high-precision ultraviolet laser cutting. As shown in Figure S3b, the multilayer film was first placed on a glass plate, after which the film was patterned with the cross-shaped trench structure using a computer-controlled laser cutter.

**Experimental set-up**

The bespoke experimental setup aims to repeatedly apply vertical forces onto the sensor with given displacement. The setup's actuation mechanism consists of a magnetic linear motor (Faulhaber LM038004001) and a 3-D printed cone-shaped probe (1 mm tip diameter). The position sensor in the linear motor provides highly accurate displacement measurements with a precision of 1 μm and is capable of outputting and recording displacement data. The probe driven by the linear motor enables applying a controllable and high-precision displacement against the sensor surface.

**Calibration of force-displacement relationship**

We placed the targeted sensing point at the centre of the bespoke probe. A high-precision scale (MAXREFDES82) was placed beneath the VBTS, which measures the forces applied. Subsequently, the position of the probe was adjusted to a height where it just contacted the sensor body. By initializing the linear motor with a pre-set displacement, we were able to measure the change of forces with the scale. We measured the force and The sensing points located at the top left corner of the sensor were tested: (1,1), (2,1), (2,2), (3,1), (3,2), and (3,3). Hence, the rest sensing points can be estimated symmetrically. For each sensing point, displacements of 0.5, 1.0, 1.5, and 1.75 mm were applied. And each force application was repeated for 10 seconds with a frequency of 0.5 Hz.

# Acknowledgments

The authors would like to acknowledge the financial support of the Engineering and Physical Sciences Research Council, UK. Grant number: EP/P012779/1. The authors would like to thank Prof Andrew Holmes for providing the laser processing equipment, and the Imperial College Research Computing Services team for promptly support in high performance computing.

# Author information

Authors and Affiliations

**Department of Electrical and Electronic Engineering, Imperial College London, London, UK**

Mayue Shi, Yongqi Zhang, Xiaotong Guo and Eric M. Yeatman

Contributions

E.M.Y. and M.S. conceived the idea and designed the experiment. M.S. completed the sensor fabrication and characterization, built the image capturing pipeline. Y.Z. completed the mathematical modelling and simulation, analysed the material mechanics data. X.G. prepared the bespoke experimental setup for displacement and force measurements. M.S. and X.G. conducted the experiments and collected the data. M.S. processed the experimental data, designed the algorithm, evaluated model performance. M.S., Y.Z., X.G., E.M.Y. wrote the manuscript.

Correspondence to Mayue Shi.

# Conflict of interest

The authors declare that they have no conflict of interest.

Supplementary Information

# High-Performance Vision-Based Tactile Sensing Enhanced by Microstructures and Lightweight CNN

Mayue Shi[1]*, Yongqi Zhang[1], Xiaotong Guo[1] and Eric M. Yeatman[1]

[1]Department of Electrical and Electronic Engineering, Imperial College London, London, UK.

*Email: m.shi16@imperial.ac.uk.

**Table of Contents**

**Supplementary Note 1 | Mathematical Model and Analytical Approach (Figure S1.1-S1.16)**

## 1.1 FLEXIBILITY METHOD TO ANALYZE FIXED-FIXED BEAMS

The flexibility method proves to be a valuable approach for analyzing statically indeterminate systems, enabling the elimination of redundant reactions within the system through successive replacements. In our specific case, we simplified the initial third-degree statically indeterminate system to a second-degree one by disregarding the axial indeterminacy and focusing on the shear and moment. Consequently, we can free one of the fixed ends, as shown in **Figure S1.1**, resulting in a degree of freedom (DOF) associated with translation (termed as DOF #1) and another associated with rotation (termed as DOF #2). This reduction in degrees of freedom streamlines the analysis and facilitates a more manageable understanding of the system's behavior.

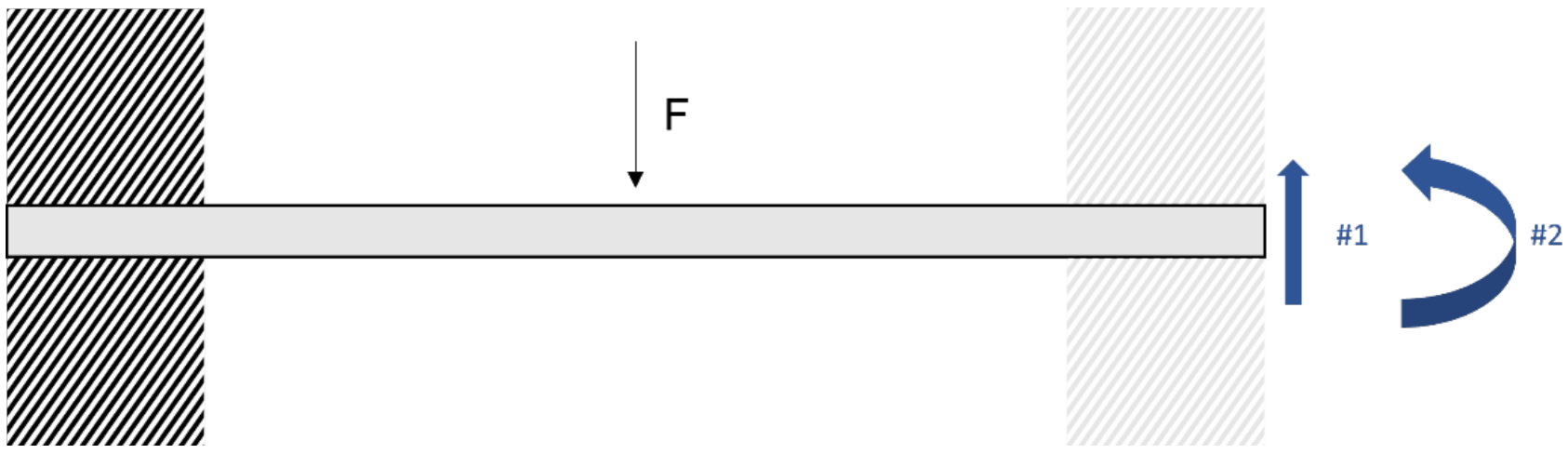

**Figure S1.1**: Using the flexibility method to analyze statically indeterminant systems. By releasing the boundary condition at one end (in this case, the right end), we are going two have two DOFs associated with translation (#1) and rotation (#2), respectively.

Given the nature of the fixed-fixed setup, the displacements in the translational direction, denoted as $\Delta_1$, and the rotational direction, denoted as $\Delta_2$, are constrained to be zero. It's important to note that we have established a convention where upward displacements are considered positive.

$$\Delta_1 = \Delta_2 = 0 \tag{S1}$$

Returning to the original system, comprising a fixed-fixed beam and a concentrated load F, we can deconstruct this system into the sum of the primary loading and the redundant reactions of a fixed-free beam, as illustrated in **Figure S1.2**.

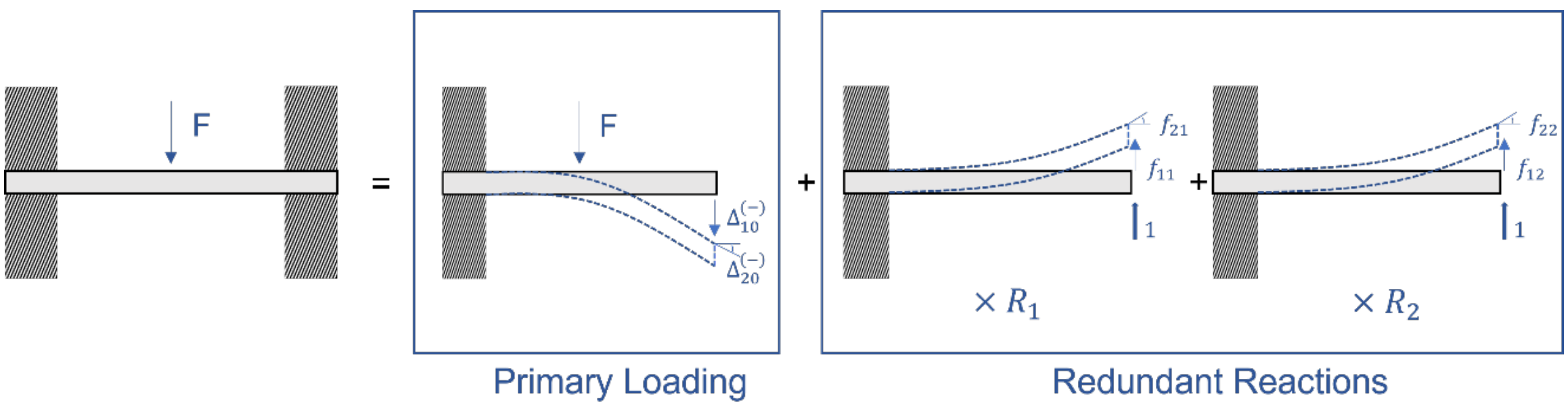

**Figure S1.2**: Full original system of a fixed-fixed beam could be dissected Primary Loading plus Redundant Reactions

The full original system of a fixed-fixed beam could be dissected as, 1. Primary loading: We disassemble the full system by removing the reaction at one end and allowing the structure to curve and displace accordingly. Translation will be denoted as $\Delta_{10}$, where the subscript 1 denotes the relevant degree of freedom (DOF #1), and 0 denotes the primary loading situation.

The same logic applies to rotation, $\Delta_{20}$. In both cases, following our convention, $\Delta_{10}$ and $\Delta_{20}$ will be negative. 2. Redundant reactions: We reintroduce the redundant reactions by multiplying them by the actual yet unknown redundant forces $R_1$ and $R_2$. For practical purposes, we use a unit load of 1 for both redundant shear and moment to trigger responses denoted by flexibility coefficients $f_{11}$, $f_{12}$, and rotation coefficients $f_{21}$, $f_{22}$.

Based on where the load is applied, we are able to formulate the translations $\Delta_{10}$ and $\Delta_{20}$, as well as flexibility coefficients $f_{11}$, $f_{12}$, and rotation coefficients $f_{21}$, $f_{22}$ in terms of load $F$, length of the beam $L$, Young's Modulus of the material $E$, and moment of inertia $I$. Then we can proceed to write out the compatibility equations to this system. For shear/translation $\Delta_1$, which ultimately is equal to 0, we have,

$$\Delta_1 = \Delta_{10} + f_{11}R_1 + f_{12}R_2 = 0 \tag{S2}$$

Similarly, for rotation $\Delta_2$, we have,

$$\Delta_2 = \Delta_{20} + f_{21}R_1 + f_{22}R_2 = 0 \tag{S3}$$

We can then put these relations in a matrix form as $\{\Delta - \Delta_0\} = [f]\{R\}$ to solve for redundant reaction forces $R_1$ and $R_2$ via calculating the determinant of the flexibility matrix $[f]$, and ultimately obtain the corresponding moment and shear plots.

**1.2 SAINT-VENANT'S PRINCIPLE IN FINITE ELEMENT METHOD**

Saint-Venant's principle, a fundamental concept in structural engineering, is a commonly utilized notion among structural engineers. Essentially, Saint-Venant's principle asserts that the precise distribution of a load becomes less critical as we move away from the loaded region, as long as the overall load resultants remain accurate. We leverage the Saint-Venant's principle, with a specific emphasis on its application within the domain of finite element (FE) analysis.

We can begin with a relatively straightforward scenario: a thin rectangular plate featuring a circular hole at a certain distance from the loaded edge, subjected to axial pulling. The main question to explore is the significance of the actual load distribution when we are particularly interested in stress concentration around the hole. In this case, we consider three distinct load types applied at the rightmost boundary:

> Scenario 1. Consistent axial stress of 100 MPa
> Scenario 2. Symmetric parabolic stress distribution with peak amplitude 150 Mpa
> Scenario 3. Centrally placed point load, engineered to yield an equivalent resultant compared to the previous two load cases

The stress distribution at the hole, as depicted in **Figure S1.3**, conspicuously showcases that the mode of load application doesn't notably influence the stress concentration around the hole. The critical factor lies in the hole being sufficiently distant from the applied load.

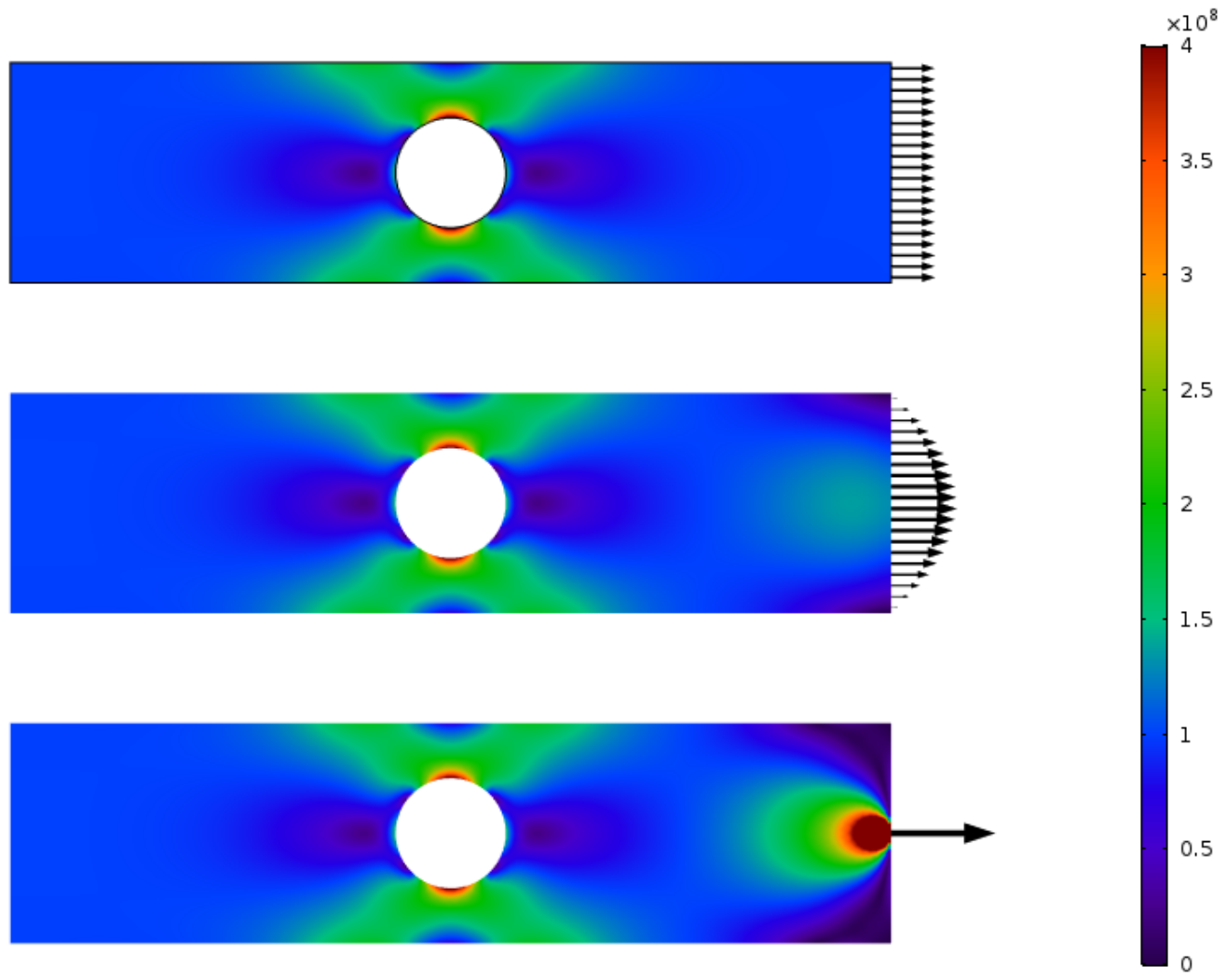

**Figure S1.3**: Von Mises stress contours for the three load cases. These contours visually demonstrate that the Von Mises stress distribution around the hole remains consistent regardless of the mode of load application. Scenario 1. Consistent axial stress of 100 Mpa, Scenario 2. Symmetric parabolic stress distribution with peak amplitude 150 Mpa, Scenario 3. Centrally placed point load, engineered to yield an equivalent resultant compared to the previous two load cases.

An alternative and insightful way to depict this scenario is through principal stress arrows. This visualization emphasizes the stress field as a flux, providing a clear representation of stress redistribution. **Figure S1.4** showcases the principal stress arrows for the three load cases. These arrows effectively illustrate stress redistribution and provide valuable insights into how stress is distributed in the structure for each load case.

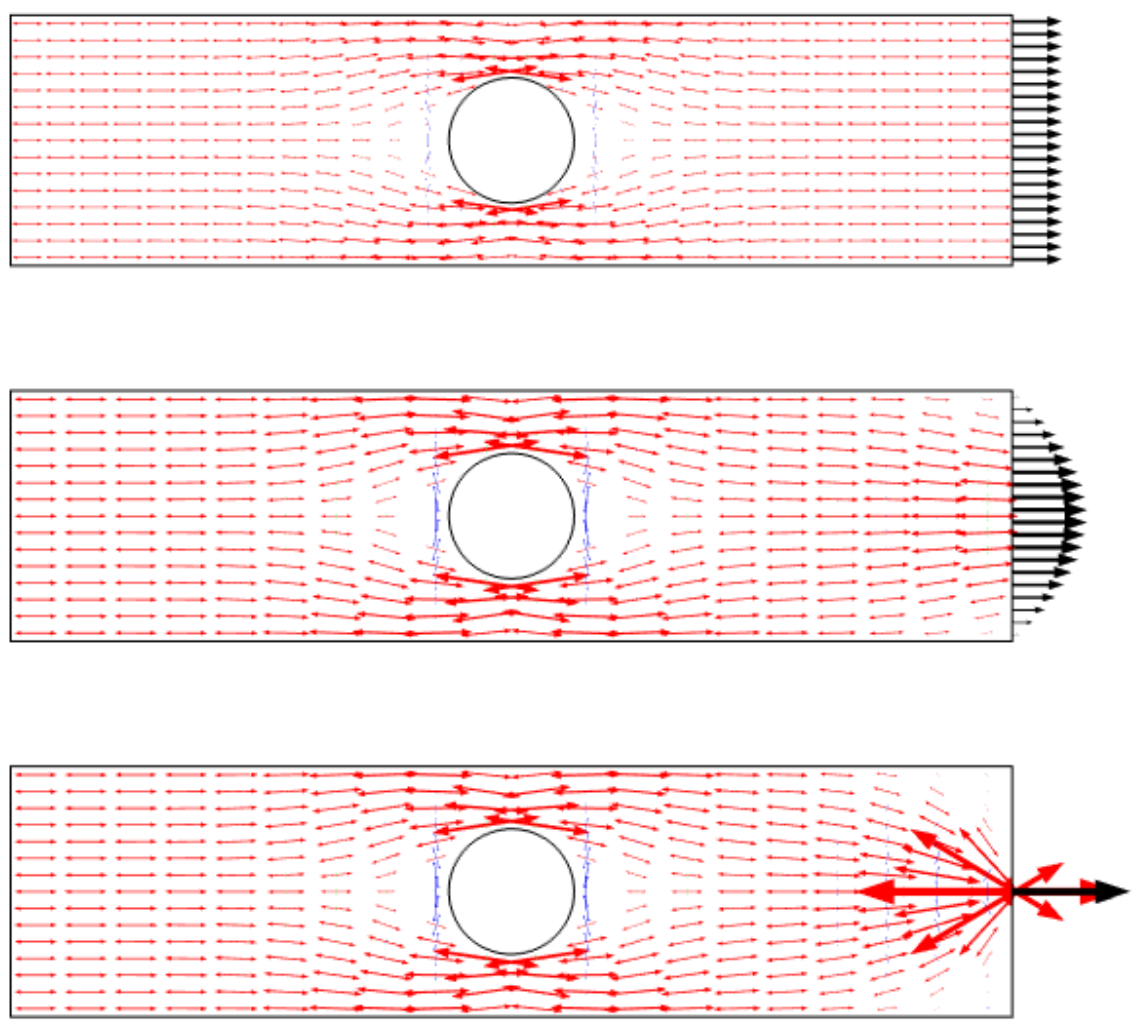

**Figure S1.4**: Principal stress plot for the three load cases (blue arrows demonstrate compression, while red arrows extension). Note a singularity when point load is used in Scenario 3.

Analyzing stress distribution along a line, as depicted in **Figure S1.5**, provides valuable insights. It becomes evident that as we move away from the loaded edge, approximately reaching the width of the plate, the stress profiles from the three cases start to align and converge. This convergence point serves as a critical observation, signifying that beyond this distance from the loaded edge, the stress distribution is no longer influenced by the distinct load types applied. Instead, it settles into a consistent pattern that remains largely unaffected by the specific way the load is applied.

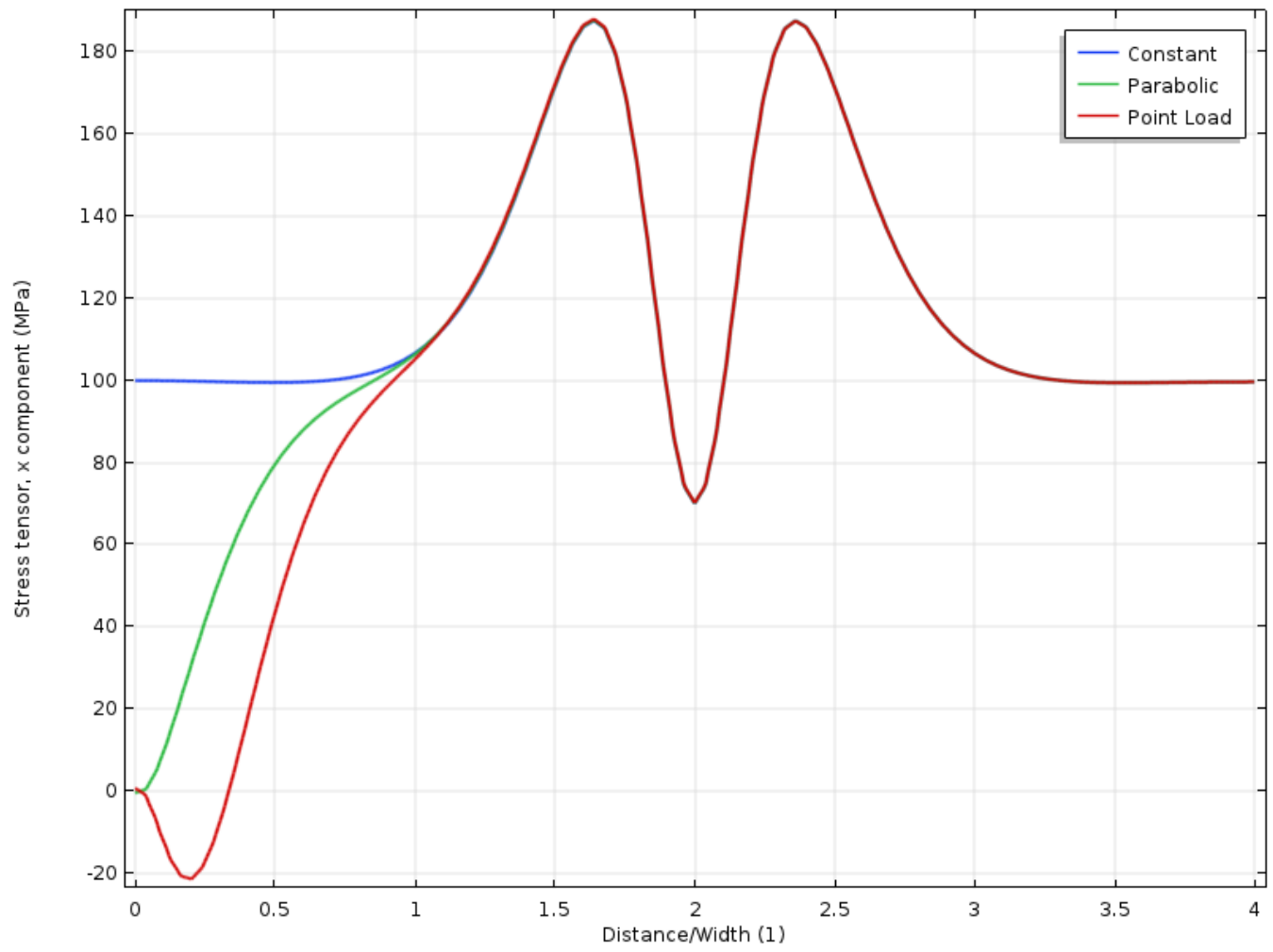

**Figure S1.5**: Stress along the upper edge, as a function of the distance from the loaded boundary. The distance is normalized by the width of the plate.

When relocating the hole closer to the loaded boundary, a noticeable transformation occurs in the stress distribution around the hole, signifying a dependency on the load distribution. Illustrated in **Figure S1.6**, the point where the stress fields from the three load cases come together is now at a distance twice as far from the loaded boundary compared to the previous scenario. This shift in the convergence distance holds significant implications—it emphasizes a fundamental principle: for Saint-Venant's principle to hold, stresses must be allowed to redistribute freely. However, the presence of the hole in this case partially hampers this redistribution, leading to the observed alterations in stress patterns.

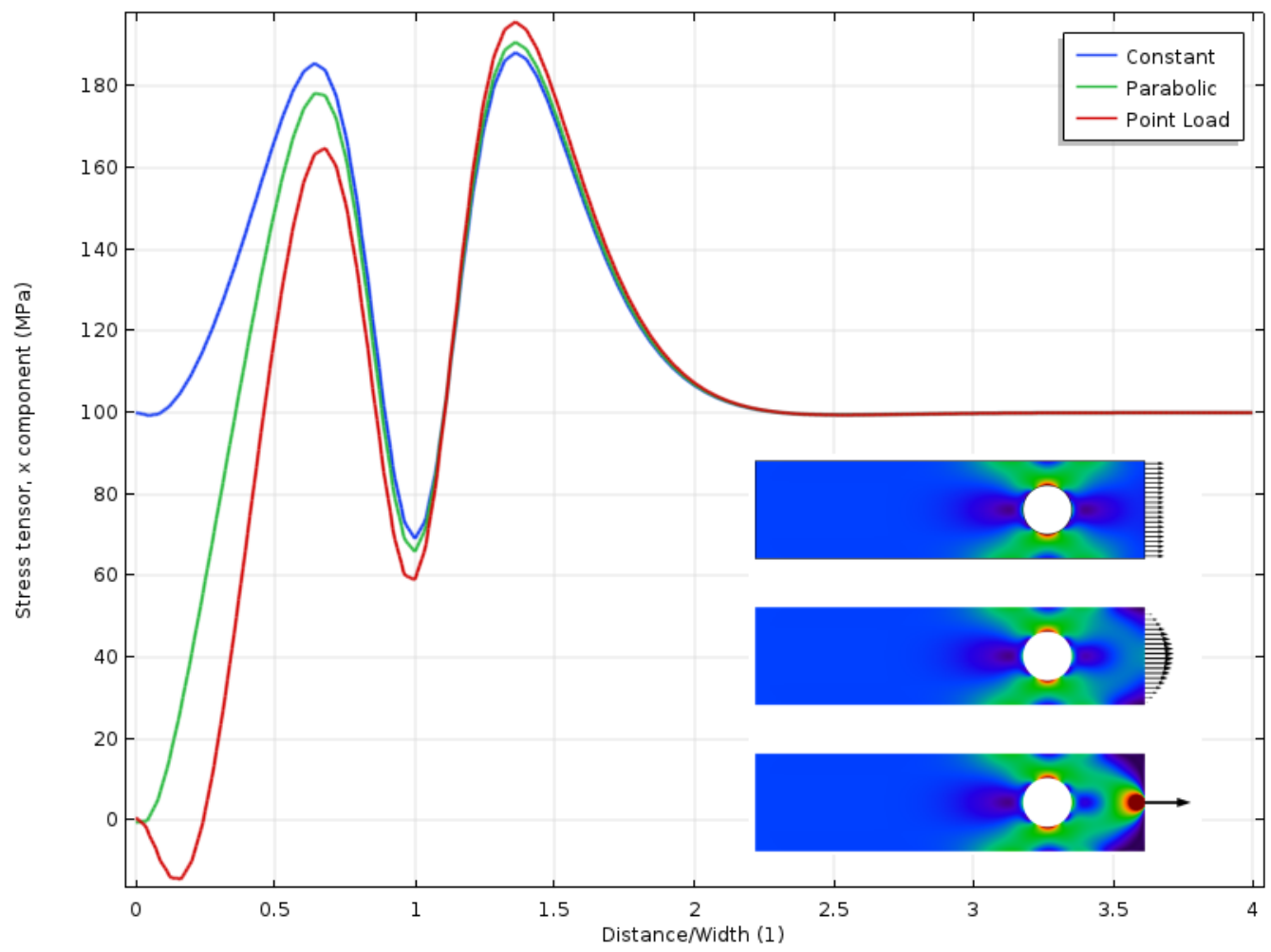

**Figure S1.6**: Stress along the upper edge with the hole closer to the loaded boundary. The distance to the point where the stress fields from the three load cases converge is now twice as far from the loaded boundary.

It's important to highlight that Saint-Venant's principle emphasizes that there is negligible difference in the stress state at a distance comparable to the linear dimension of the loaded area. However, a key consideration is that the loaded area to be considered may not necessarily be the actual area that is physically loaded. When the hole is situated far away, we often determine stress concentration factors using handbooks. The handbook approach implicitly assumes even load distribution, similar to the first load case. Consequently, even if the actual load is applied to a small portion of the boundary, the critical distance is related to the size of the entire boundary.

In contrast, when employing the finite element method (FEM), the hole can be placed very close to the load. The limiting factor is the well-defined load distribution from a physical standpoint. However, it's crucial to acknowledge that assumptions about redistribution implicitly assume a specific load distribution, which may not align with the actual distribution. This distinction underscores the nuanced nature of stress analysis and the implications of assumptions on load distribution in different methodologies.

## 1.3 DEFORMATION MODELING OF LOADING UNNOTCHED BEAM

As introduced, upon simplification, the system transforms into a beam with two fixed ends, as illustrated in **Figure S1.7**. This configuration introduces three potential reactions at each end, rendering the system statically indeterminate to the third degree. It's noteworthy that one of these reactions operates exclusively in the axial direction and remains uncoupled from both shear and moment. To streamline the analysis practically, we can disregard this axial reaction, consequently simplifying the system to a second-degree static indeterminacy.

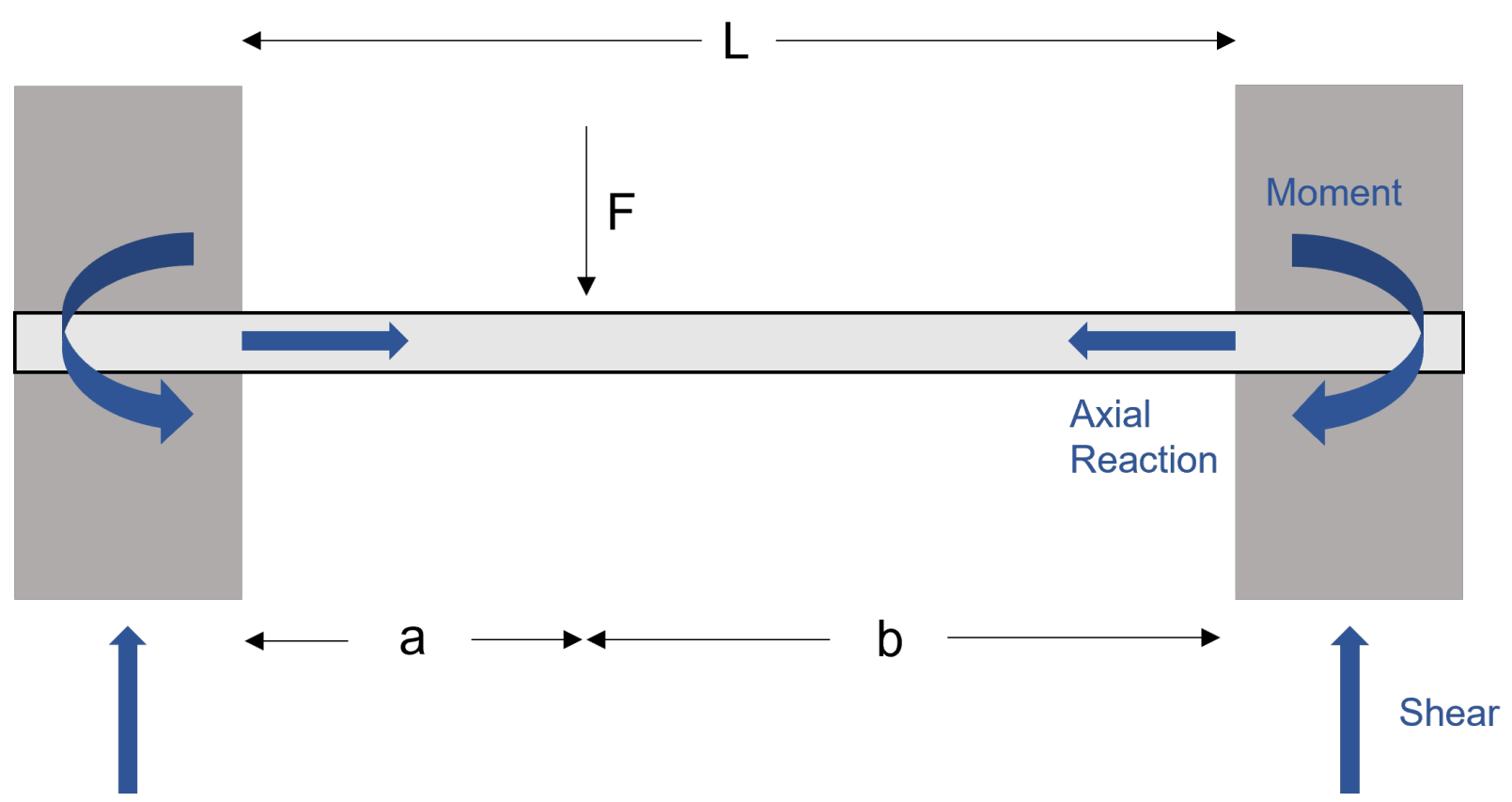

**Figure S1.7**: The reactions to bending of a generic fixed-fixed beam. In this illustration, a beam of length L is fixed at both ends, and a concentrated load F applied at an arbitrary location induces three reactions, namely axial, shear, and moment. However, for pragmatic considerations, we concentrate solely on analyzing the shear and moment reactions, as the axial reaction remains decoupled from these responses.

This simplification allows us to focus solely on shear and moment reactions for a more targeted analysis. Additionally, we proceed with the assumption that the beam exhibits linear elastic behavior, maintaining a uniform Young's modulus (E) and moment of inertia (I) throughout. This assumption is fundamental to our analytical approach and provides a solid basis for modeling the beam's deformation patterns accurately.

The flexibility method serves as a powerful tool for analyzing statically indeterminate systems. This approach involves eliminating redundant reactions within the system and replacing them with appropriate degrees of freedom (DOFs), enabling a systematic analysis of the structure. By applying the flexibility method systematically, we can construct a detailed moment diagram illustrating the bending of a fixed-fixed beam. This comprehensive diagram accounts for the loads applied at notched locations labeled as A, B, C, D, and E, each specified in **Figure S1.8**.

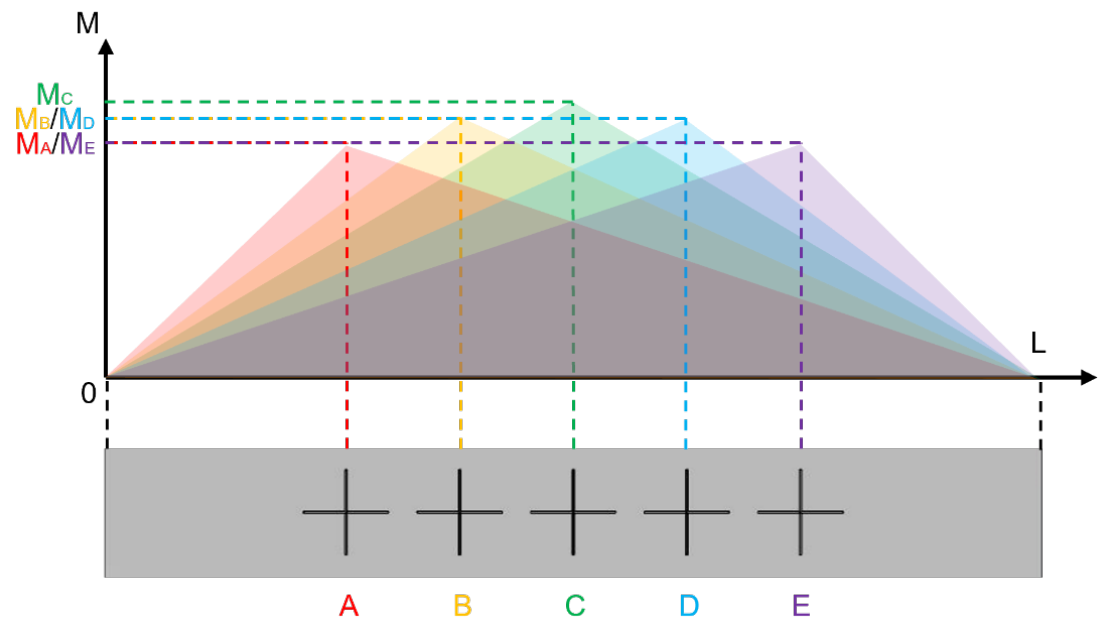

**Figure S1.8**: Comprehensive moment diagram of the system, taking into consideration the loads applied at notched locations A, B, C, D, and E. The moments for each location are denoted correspondingly as $M_A$, $M_B$, $M_C$, $M_D$, and $M_E$.

The maximum moment, $M_{max}$, at each location is calculated using $M_{max} = \frac{Fab}{L}$, where $F$ is the applied load, $L$ is the length of the beam, and $a$ and $b$ are distances from the point of application to the two ends of the beam. Notably, the maximum moment occurs at the central location $C$, where $M_C = \frac{FL}{4}$ due to the symmetric arrangement of notches. Moreover, owing to the symmetrical layout of the notches, we can observe that $M_A = M_E$ and $M_B = M_D$, further simplify the analysis and reduce the number of calculations needed.

We began the simplification of the thin film by conceptualizing it as an array of fixed-fixed beams. In this representation, during bending, the beams do not exert pressure on each other, assuming minimal deformation of their cross-sections. Therefore, when an external load is applied to a specific region of the surface, the film undergoes bending. This bending effect can be accurately understood by considering that each beam within the array responds independently, aligning with the distinct areas of the applied load.

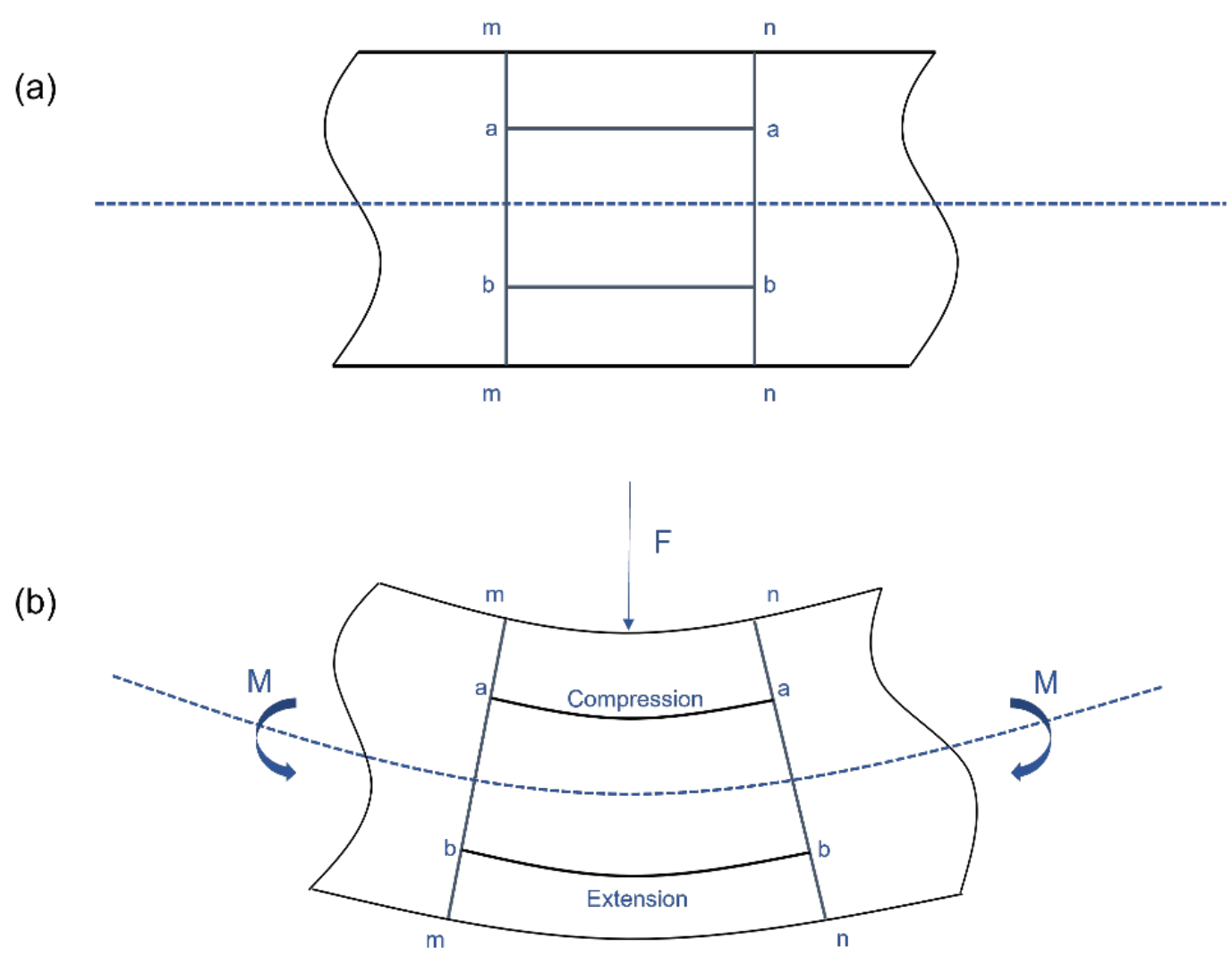

**Figure S1.9**: (a) The beam in its initial undeformed configuration (b)Subsequent deformation of the beam. The dotted line delineates the neutral layer.

In constructing a mathematical model to describe the bending behavior of this structure, our main objective revolves around understanding the relationship between the applied load and the resulting deformation, with a specific focus on the notch width. A significant advantage in analyzing beams is the ability to establish a neutral layer within each beam with uniform material distribution. During the elastic bending of the beam, this neutral layer maintains constant dimensions. Meanwhile, the top and bottom surface layers undergo extension and compression, respectively, as illustrated in **Figure S1.9**.

Leveraging the inherent properties of the neutral layer, we can readily derive numerous valuable geometric relationships. When considering an arbitrary section of the beam, we define a segment $\overline{bb}$ situated at a distance y below the neutral layer $\overline{OO}$. Both the neutral layer and segment have an original length of $dx$ before bending. We can express this relationship as follows,

$$\overline{bb} = dx = \overline{OO} \tag{S4}$$

When the beam deforms under bending, the length of the neutral layer $\overline{OO}$ remains unchanged based on its properties,

$$dx = \overline{OO} = \breve{OO} = \rho d\theta \tag{S5}$$

Where $\rho$ is the bent beam's radius of curvature with angle $d\theta$. Then, for the case of the interested segment $\overline{bb}$, we have

$$\breve{bb} = (\rho + y)d\theta \tag{S6}$$

Here, $\breve{bb}$ and $\breve{OO}$ represent the length of the segment $\overline{bb}$ and $\overline{OO}$ respectively. When the material is bent under compression. Consequently, the strain ($\epsilon$) can be defined as follows,

$$\epsilon = \frac{(\rho + y)d\theta - \rho d\theta}{\rho d\theta} = \frac{y}{\rho} \tag{S7}$$

Expanding on the geometric relationships we have established, let's further explore the fundamental physical connections between stress $\sigma$ and strain $\epsilon$. These relationships are crucial foundations for comprehending the structural behavior of materials when subjected to bending. Understanding how stress and strain interrelate is fundamental in analyzing and predicting the behavior of materials and structures under various mechanical conditions, particularly bending.

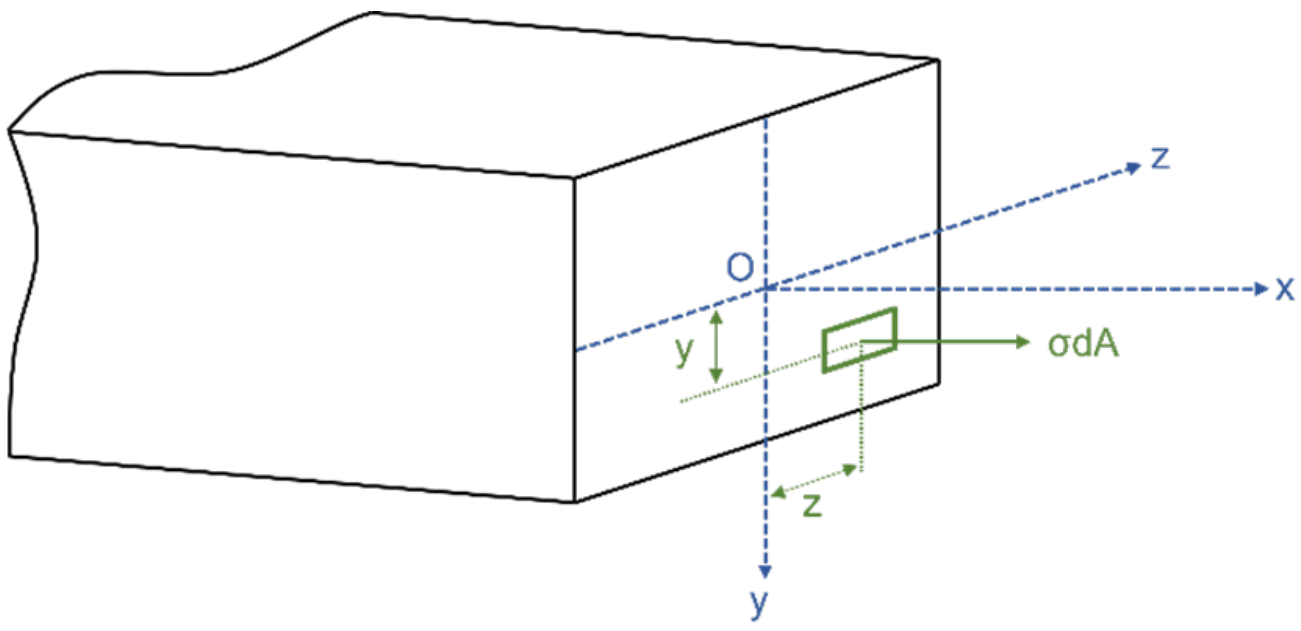

**Figure S1.10**: Examining the cross-section of the beam, we set the stage for formulating physical relationships by establishing a Cartesian coordinate system centered within the beam's cross-section. Within this framework, we consider an infinitesimal area $dA$.

We start by defining a Cartesian coordinate system, as shown in **Figure S1.10**, using the neutral layer as the reference, within the cross-section of the beam. The origin is placed at the center of the cross-section. Considering an infinitesimal area $dA$, we determine its offsets to the $y$ and $z$ axes as $y$ and $z$ respectively. With this foundation, we proceed to derive fundamental relationships that illuminate the connections between stress, strain, and the geometry of the beam. These relationships are essential in understanding how stress and strain manifest in the structure in relation to its geometry.

$$\sigma = E\epsilon \tag{S8}$$

$$\Rightarrow \sigma = E\frac{y}{\rho} \propto y \tag{S9}$$

$$\sigma = \frac{My}{I_z} \tag{S10}$$

In the above expressions, $E$ represents the Young's Modulus of the beam, $\rho$ stands for the radius of curvature of the deformation, and $I_z$ defines the 2nd Moment of Inertia (area moment of inertia) at the beam's cross-section.

To prepare for modeling the deformation of a notched beam, it's crucial to initially understand the deformation characteristics of an unnotched beam. Let's define an arbitrary beam with a length $L$, width $w$, and height $h$, as shown in **Figure S1.11**.

This unnotched beam acts as a fundamental reference, providing insights into the behavior of beams without notches. Later, we can contrast this with notched beams to comprehend the influence of notches on deformation properties.

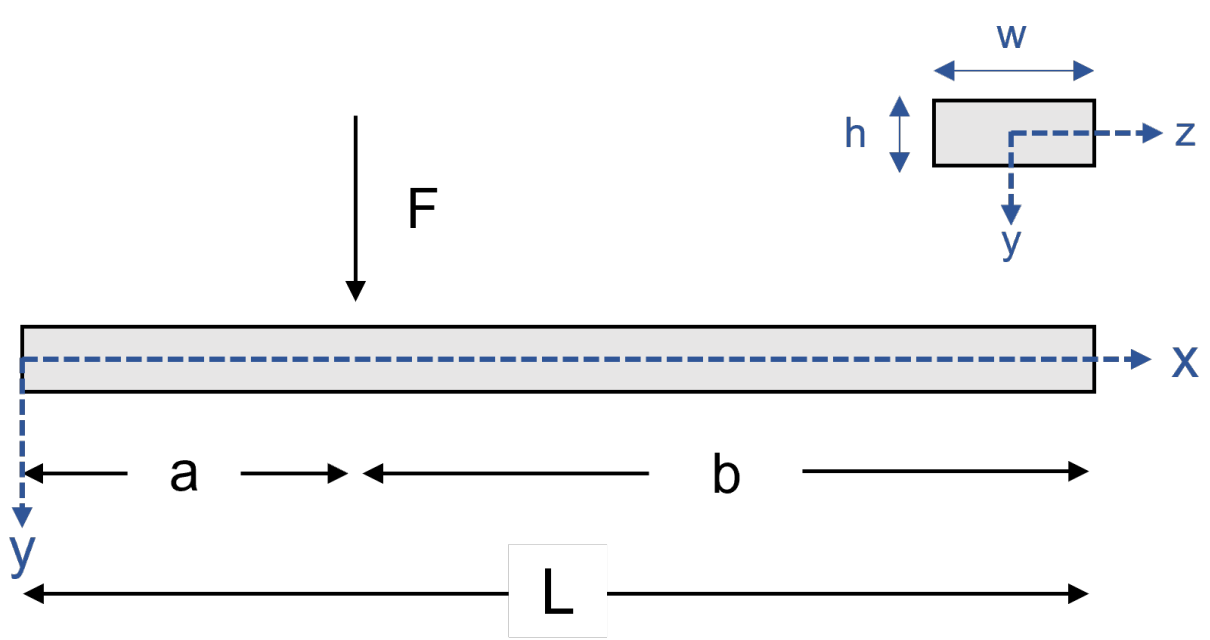

**Figure S1.11**: 3D beam of length L, width w, and height h under applied load F. A Cartesian coordinate system is attached as shown.

When a load $F$ is applied, the beam undergoes deformation. If we select an arbitrary segment $d_0$ on the opposite side of where the load is applied, it will extend by $\Delta d_0$, as illustrated in **Figure S1.12**.

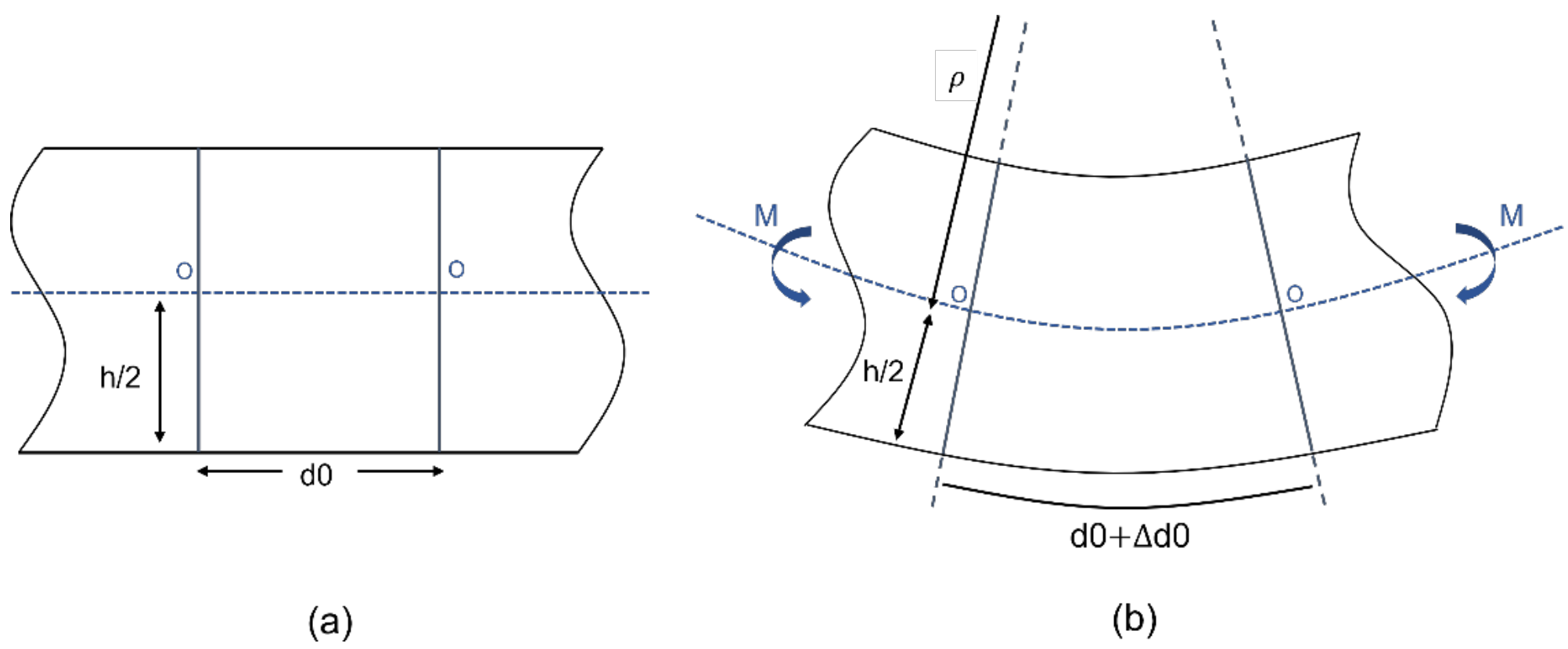

**Figure S1.12**: (a) Unnotched beam in its undeformed shape with projected notch width $d_0$. (b) Unnotched beam undergoing bending with projected deformed notch width $d_0 + \Delta d_0$.

Referring back to the geometric and physical relationships introduced in the previous section, we can establish the following relationships:

$$\overbrace{d_0 + \Delta d_0} = \left(\rho + \frac{h}{2}\right) d\theta \tag{S11}$$

$$d_0 = \overline{OO} = \overset{\frown}{OO} = \rho d\theta \tag{S12}$$

From these, we can calculate the strain in this case be,

$$\epsilon_0 = \frac{(d_0 + \Delta d_0) - d_0}{d_0} = \frac{\Delta d_0}{d_0} = \frac{h}{2\rho} \tag{S13}$$

Finally, we can proceed to calculate the deformation (in this case, extension) of $d_0$ using the following expressions:

$$\frac{6M}{wEh^2} = \frac{\Delta d_0}{d_0} \tag{S14}$$

$$\Rightarrow \Delta d_0 = \frac{6Md_0}{wEh^2} \tag{S15}$$

**1.4 DEFORMATION MODELING OF LOADING NOTCHED BEAM**

Saint-Venant's principle provides valuable insights by indicating that the stress state remains nearly unchanged at a distance approximately equal to the linear dimension of the loaded area. In cases where a disturbance, such as holes or notches, is located far from the loaded region, stress computation assumes an even load distribution. The critical distance is associated with the size of the entire boundary if the load is confined. In the context of finite element method (FEM) analysis, when the disturbance is near the load, its influence is limited by the requirement for a well-defined load distribution based on appropriate assumptions of redistribution.

In our specific scenario, we can apply Saint-Venant's principle to infer that stresses on a boundary considerably distant from the applied load remain relatively unaffected. The alterations in stress and strain primarily occur in the vicinity of the load application regions. Given that the notches are positioned on the opposite side from where the load is applied and considering the thinness of the notches relative to the beam's thickness, we can confidently employ Saint-Venant's Principle in our analysis.

To simplify our analysis leveraging Saint-Venant's principle, we opt to consider the regions immediately surrounding the notch as stress-free areas. We can start from the central location (Location C), which proves to be the most convenient for analysis due to its symmetrical properties.

Here, we represent the stress-free area as an isosceles triangle with a height of $\alpha$ and a base length of $2\alpha$, as shown in **Figure S1.13**. This approach simplifies computations and aligns with the Saint-Venant's principle's application.

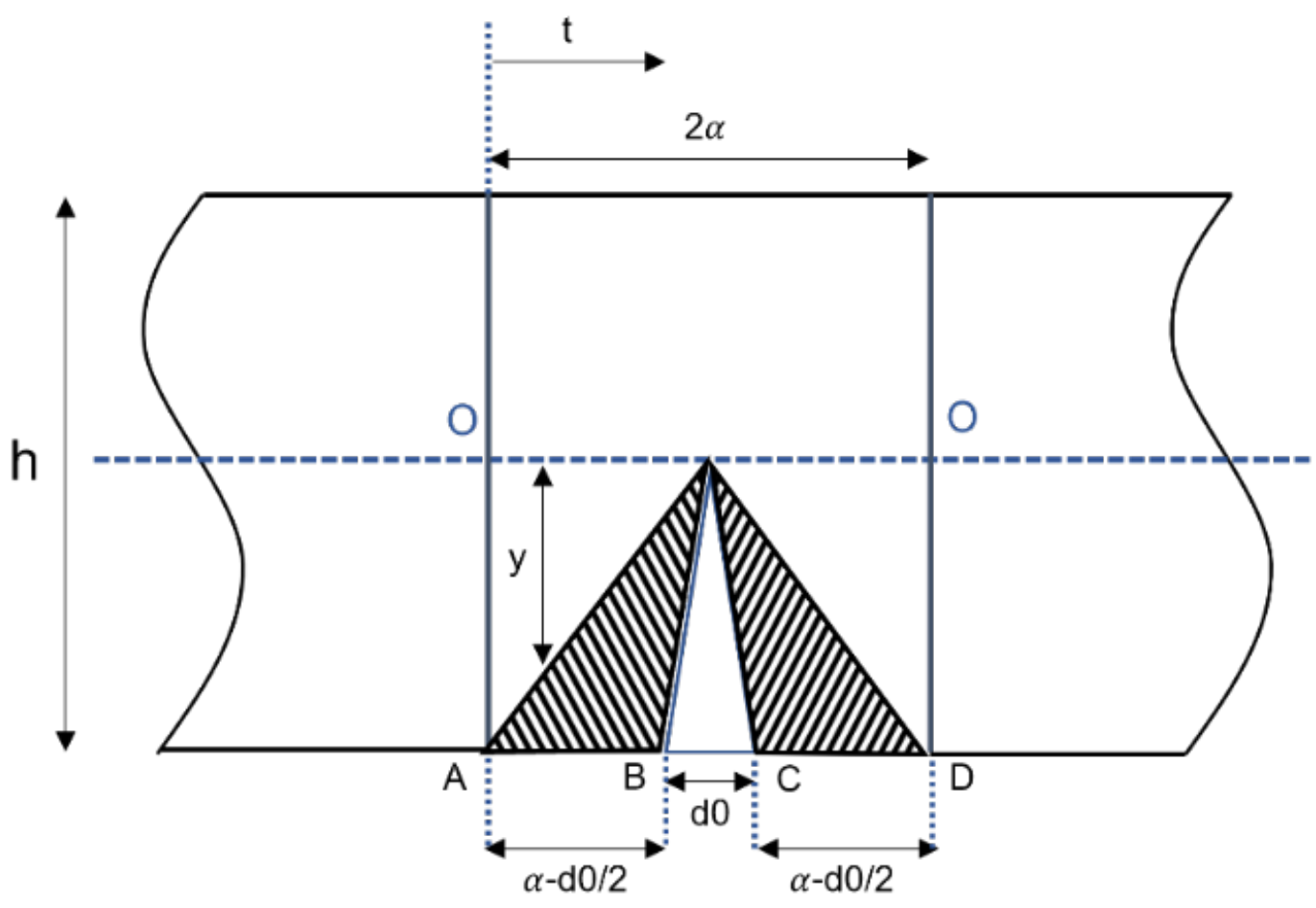

**Figure S1.13**: Simplifying the analysis of the notched beam utilizing Saint-Venant's Principle at the central location.

In this context, $d_0$ represents the original notch width situated at the center of the expanded configuration. According to Saint-Venant's principle, the expanded sections (shaded areas) will experience negligible deformation during the beam's bending. Consequently, the extension in the expanded configuration (segment $\overline{AD}$) is equal to the extension in the original configuration (segment $\overline{BC}$), denoted as $\Delta d$.

Extending from the geometric and physical relationships elucidated in previous sections, we now proceed to derive the following expressions to delineate this relationship:

$$y = \alpha - t \tag{S16}$$

$$\Rightarrow \epsilon_1 E = \frac{M_C y}{I_z} = \frac{M_C(\alpha - t)}{\frac{1}{12} wE(2\alpha - t)^3} \tag{S17}$$

$$\Rightarrow \epsilon_1 = \frac{12M_C(\alpha - t)}{wE(2\alpha - t)^3} \tag{S18}$$

Utilizing the fundamental definition of strains as the ratio of deformation to the original length, we apply basic principles of calculus to express it in terms of an arbitrary length $t$.

$$\epsilon = \lim_{t \to 0}(\Delta \text{t}) \tag{S19}$$

Through integration across the entire expanded area depicted in **Figure S1.14**, resembling an isosceles triangle, we can determine the deformation of the notch width. This integration process is crucial for quantifying the extent of deformation experienced by the notch width as the beam undergoes bending.

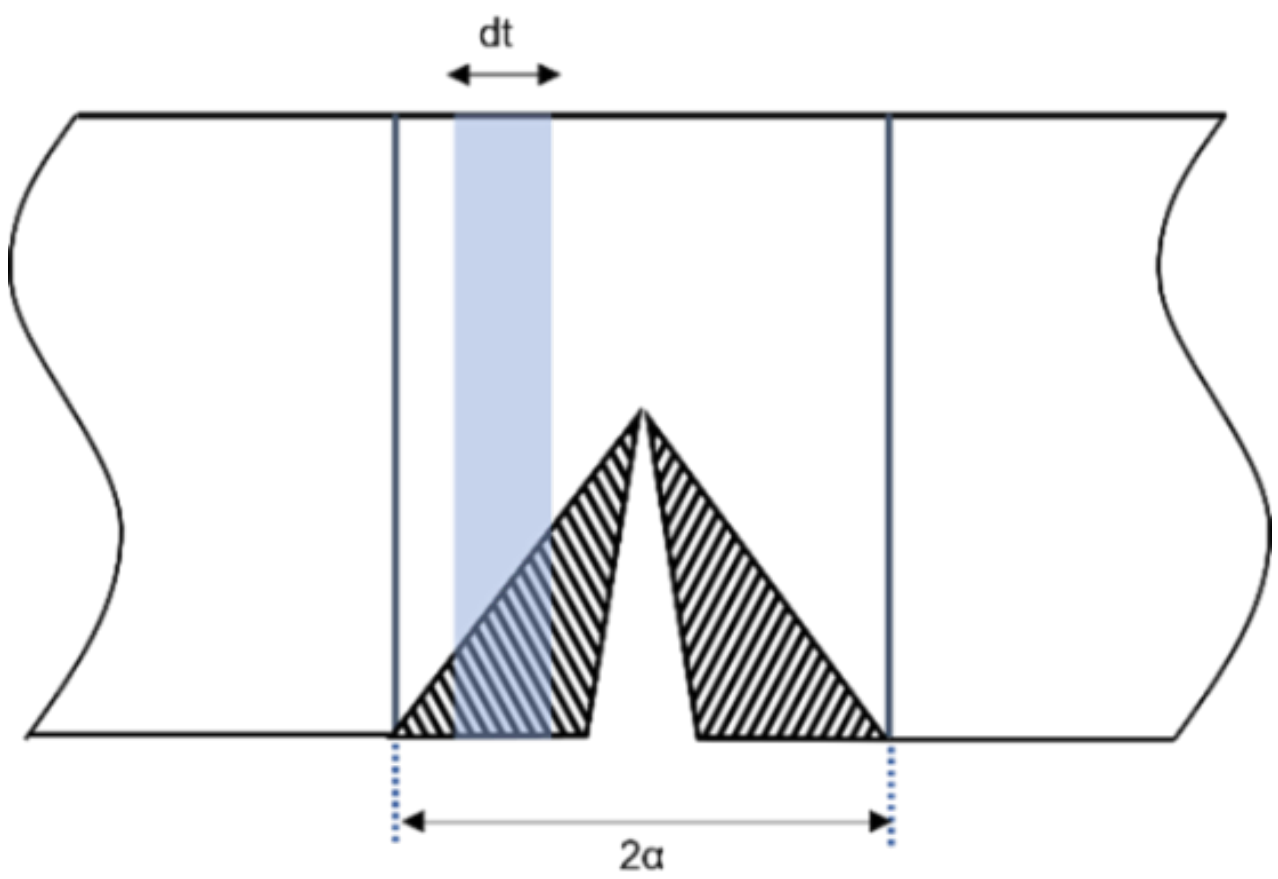

**Figure S1.14**: When integrating over the entire expanded area, represented by an isosceles triangle, we can obtain the deformation of the notch width. This integration process allows us to quantify the extent of deformation experienced by the notch width during the beam's bending. Furthermore, to ensure smooth integration and avoid discontinuity at the tip of the triangle, we consider integration over a symmetrical area. This allows us to perform the following operations to derive the deformation:

$$\Delta d_C = \int_0^{2\alpha} \epsilon_1 dt = 2\int_0^{\alpha} \epsilon_1 dt \tag{S20}$$

$$\Rightarrow \Delta d_C = \frac{3M_C}{wE\alpha} \tag{S21}$$

When analyzing intermediate locations (A, B, D, and E) that are off-center, a notable asymmetry becomes evident regarding the stress-free areas. These stress-free areas closely resemble the shape of the respective moment diagrams at each location.

Saint-Venant's principle indicates that the stress-free area is inversely related to deformation, where smaller stress-free areas experience more significant deformation and vice versa. Therefore, during integration, similar to the central location, we should integrate using a horizontal mirror image of the corresponding moment diagram to accurately capture the varying deformation across these off-central locations. This observation emphasizes the importance of considering the shape of stress-free areas in accurately analyzing and interpreting the beam's behavior at different positions.

We initiate our analysis with intermediate location A (with similar considerations applicable to location E due to symmetry), focusing on the geometric representation of the stress-free area. Here, we model the stress-free area's geometry as a horizontal mirror image of the moment diagram. This stress-free area takes the form of a scalene triangle characterized by a height of $\alpha$ and a base length of $2\alpha$, as depicted in **Figure S1.15** through the shaded regions.

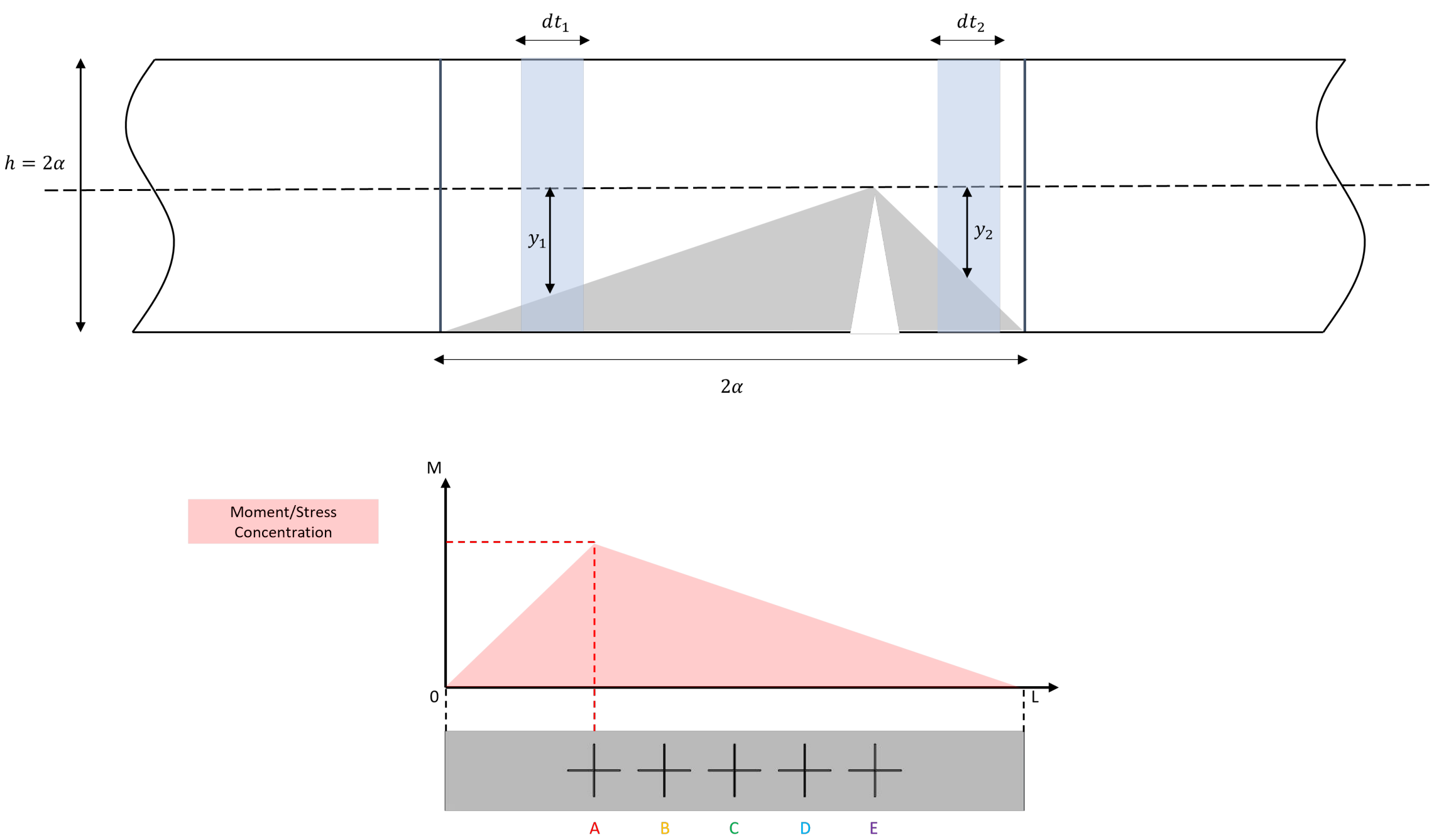

**Figure S1.15**: Representation of the stress-free area at intermediate location A. Unlike the central location, the stress-free area is not an isosceles triangle. Instead, it is shaped in accordance with the corresponding moment diagram. The stress-free area's geometry reflects the moment diagram, showcasing a shape that accurately captures the stress-free region for this location.

Expanding on the previously established geometric and physical relationships, we now derive expressions that encapsulate this relationship, emphasizing the introduction of two parameters, $t_1$ and $t_2$. This represents a departure from the formulation for the central location C, which involved only a single parameter. Additionally, we incorporate the specific geometry of the notch into our analysis. Location A is positioned at 4.25mm from the closer boundary, relative to the entire beam, which has a length of 16.5mm. Consequently, this gives the moment $M_A = \frac{833}{4356}FL$.

$$y_1 = \alpha - \frac{165}{245}t_1 \quad y_2 = \frac{165}{85}t_2 \tag{S22}$$

$$\epsilon_{A1} = \frac{M_A y_1}{I_z E} = \frac{12M_A(\alpha - \frac{165t_1}{245})}{wE\left(2\alpha - \frac{165t_1}{245}\right)^3} \tag{S23}$$

$$\epsilon_{A2} = \frac{M_A y_2}{I_z E} = \frac{12M_A(\frac{165t_1}{85})}{wE\left(\alpha + \frac{165t_1}{85}\right)^3} \tag{S24}$$

By adhering to strain's fundamental definition as deformation relative to the original length and utilizing principles of calculus, we perform integration across the shaded area in **Figure S1.15** to calculate the deformation of the notch width at this intermediate location. To handle the discontinuity at the triangle's peak, we conduct two separate integrations. These separate integrations are vital for accurately calculating deformations at the left and right edges of the notch.

$$\Delta d_A = \int_0^{\frac{245\alpha}{165}} \epsilon_{A_1} dt + \int_0^{\frac{85\alpha}{165}} \epsilon_{A_2} dt \tag{S25}$$

$$\Delta d_A = \Delta d_{A_1} + \Delta d_{A_2} = \frac{49M_A}{22wE\alpha} + \frac{17M_A}{22wE\alpha} = \frac{3M_C}{wE\alpha} \tag{S26}$$

Lastly, shifting our focus to intermediate location B (with analogous considerations applicable to location D due to symmetry), we employ a method similar to our approach for formulating deformation at intermediate locations A and E. Here, we model the stress-free area's geometry as a horizontal mirror image of the moment diagram. This stress-free area takes on the configuration of a scalene triangle, characterized by a height of $\alpha$ and a base length of $2\alpha$, as depicted in **Figure S1.16** through the shaded regions.

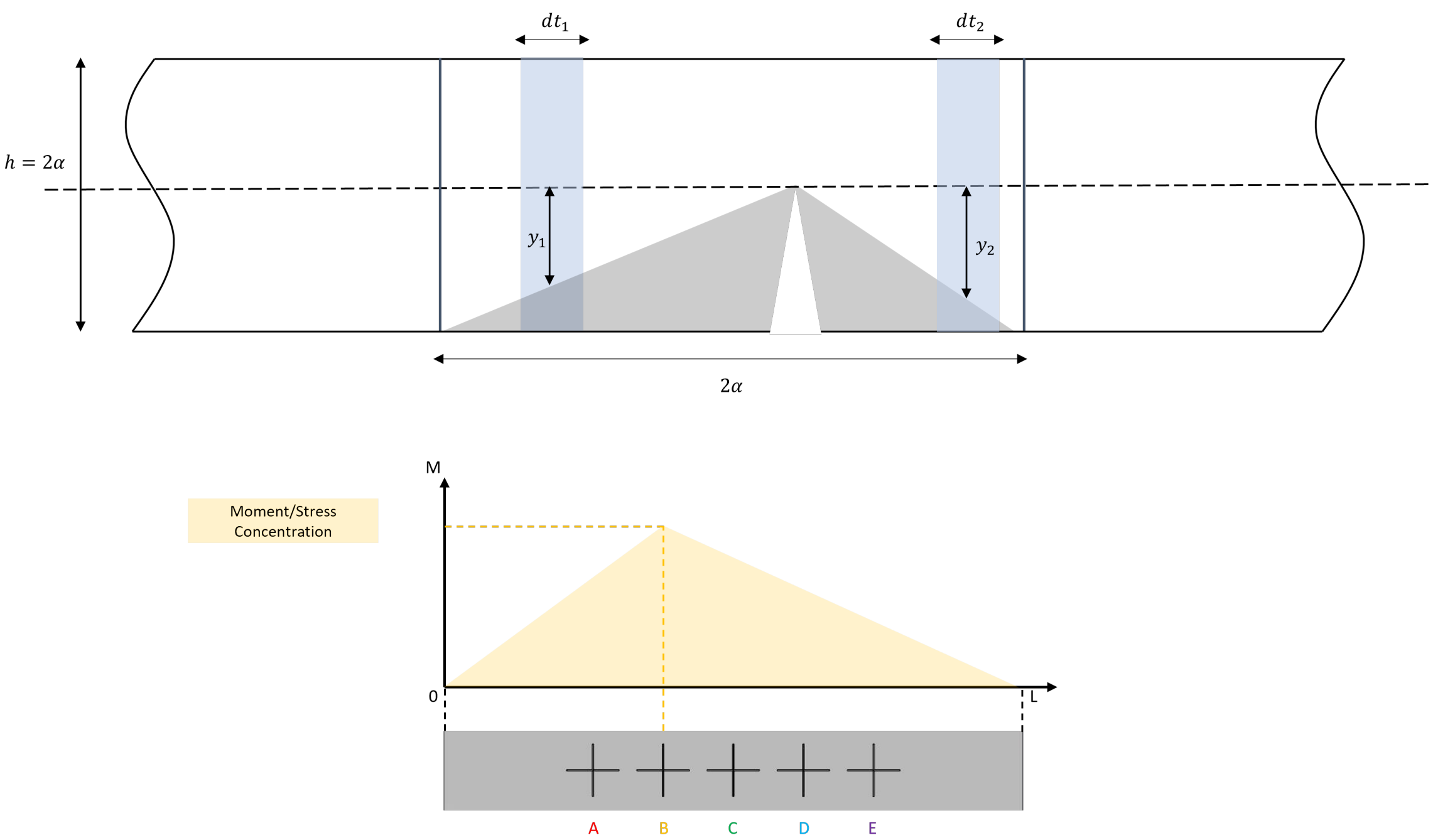

**Figure S1.16**: Graphical representation of the stress-free area at intermediate location B. This stress-free area conforms to the shape dictated by the corresponding moment diagram, mirroring its configuration. The illustration vividly portrays the accurate delineation of the stress-free region specific to this location, effectively capturing the structural behavior in this region.

Building upon established geometric and physical relationships, we proceed to derive expressions that encapsulate this relationship. Once again, we introduce two parameters, $t_1$ and $t_2$, while assimilating the specific geometry of the notch into our analytical framework. Notably, location B is positioned at 5.75mm from the nearer boundary, within the context of the beam that spans a length of 16.5mm. Consequently, this determination yields $M_B = \frac{989}{4356}FL$.

$$y_1 = \alpha - \frac{165}{215}t_1 \quad y_2 = \frac{165}{115}t_2 \tag{S27}$$

$$\epsilon_{B_1} = \frac{M_B y_1}{I_z E} = \frac{12M_B(\alpha - \frac{165t_1}{215})}{wE\left(2\alpha - \frac{165t_1}{215}\right)^3} \tag{S28}$$

$$\epsilon_{B_2} = \frac{M_B y_2}{I_z E} = \frac{12M_B(\frac{165t_1}{115})}{wE\left(\alpha + \frac{165t_1}{115}\right)^3} \tag{S29}$$

Following the same procedure, we perform integration over the shaded region in **Figure S1.16**. To address the discontinuity at the triangle's peak, we conduct two separate integrations for deformation calculations at the left and right edges. This approach ensures accurate deformation calculations for both sides of the notch.

$$\Delta d_B = \int_0^{\frac{215\alpha}{165}} \epsilon_{B_1} dt + \int_0^{\frac{115\alpha}{165}} \epsilon_{B_2} dt \tag{S30}$$

$$\Delta d_B = \Delta d_{B_1} + \Delta d_{B_2} = \frac{43M_B}{22wE\alpha} + \frac{23M_B}{22wE\alpha} = \frac{3M_C}{wE\alpha} \tag{S31}$$

## Supplementary Figure S2

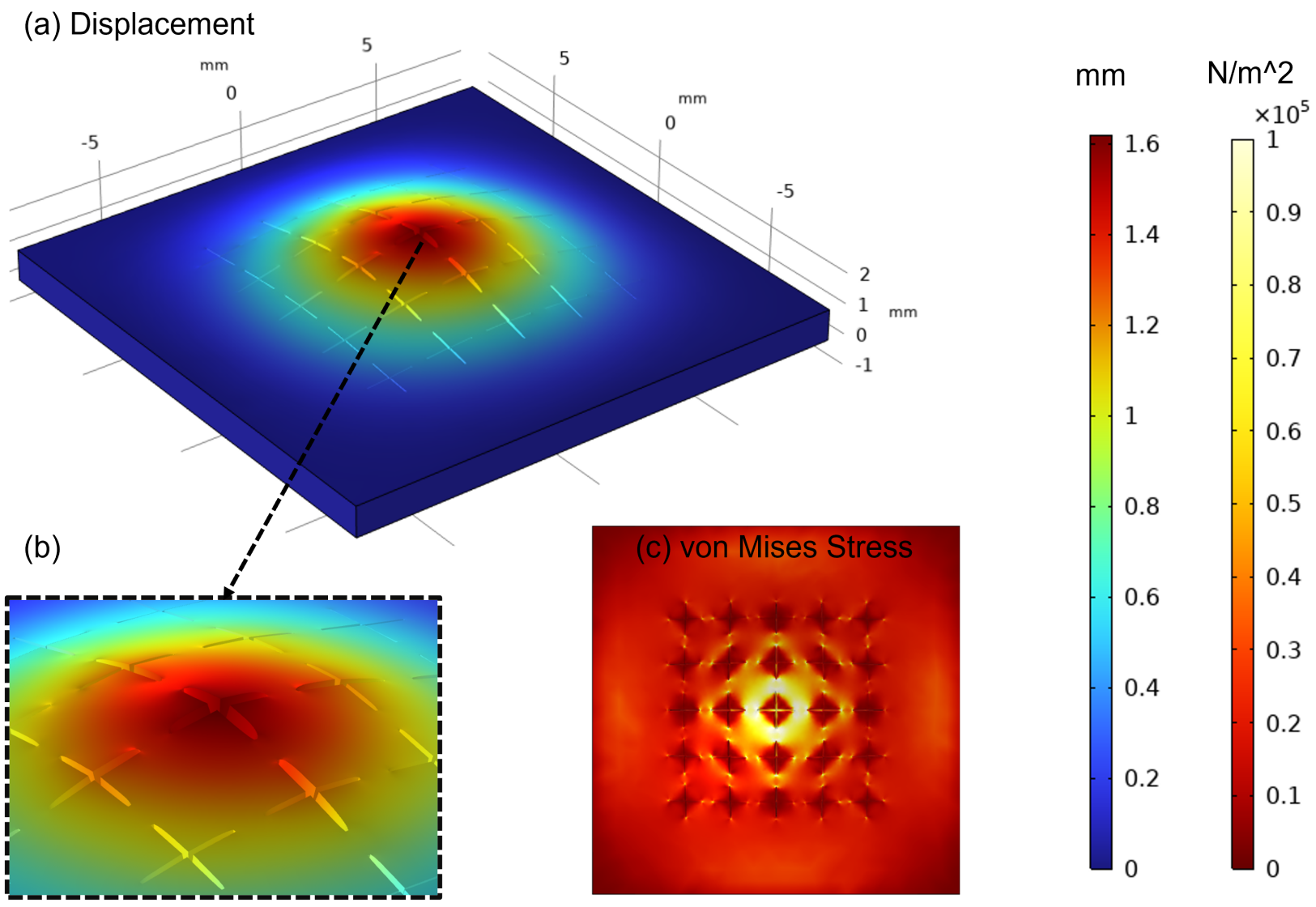

**Figure S2**: Finite element analysis of (a) deformation and (b) stress distribution with a force of 60 mN applied at the bottom centre (3, 3) of the sensor along the y-direction (diameter of contact area: 1 mm).

**Supplementary Figure S3**

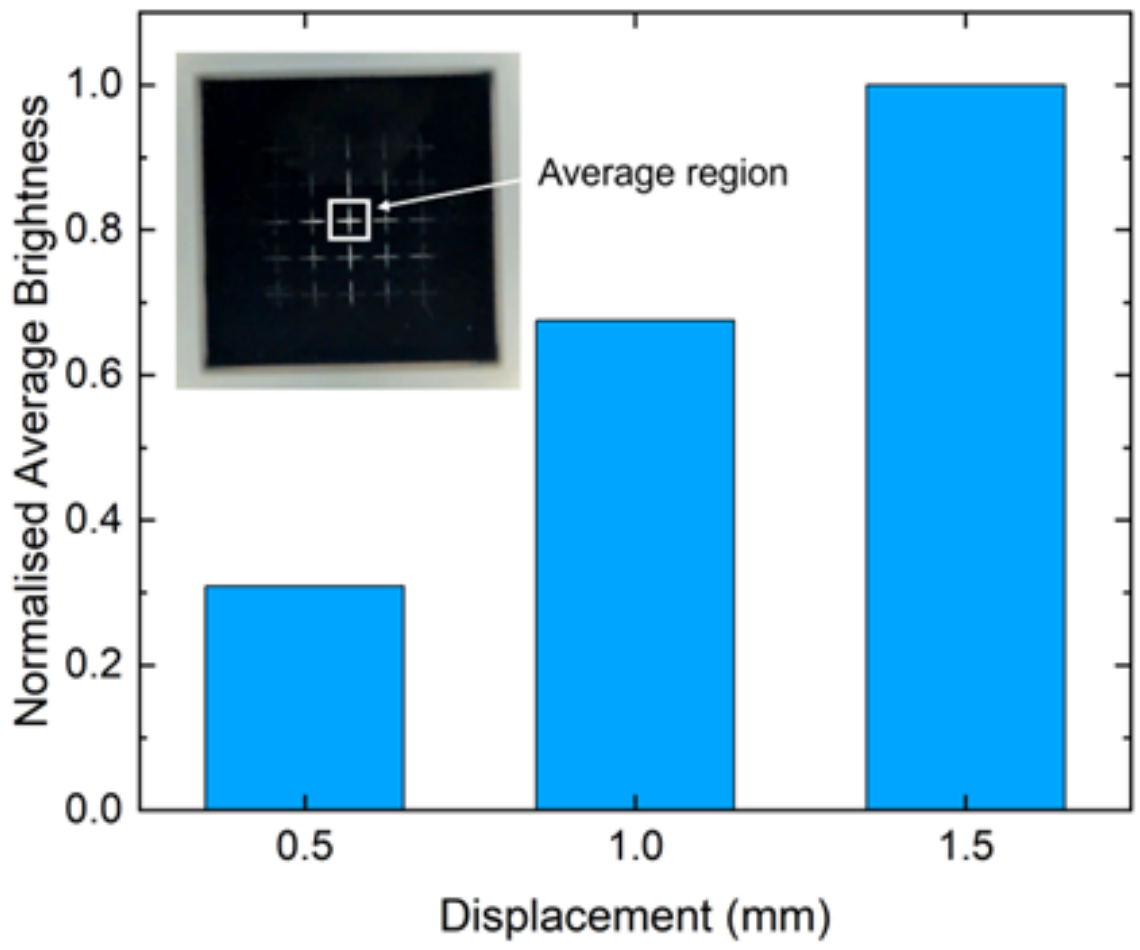

**Figure S3**: Variation of average brightness with different displacements. The contact point was at (3,3), the square average region fully included the central cross.

**Supplementary Figure S4**

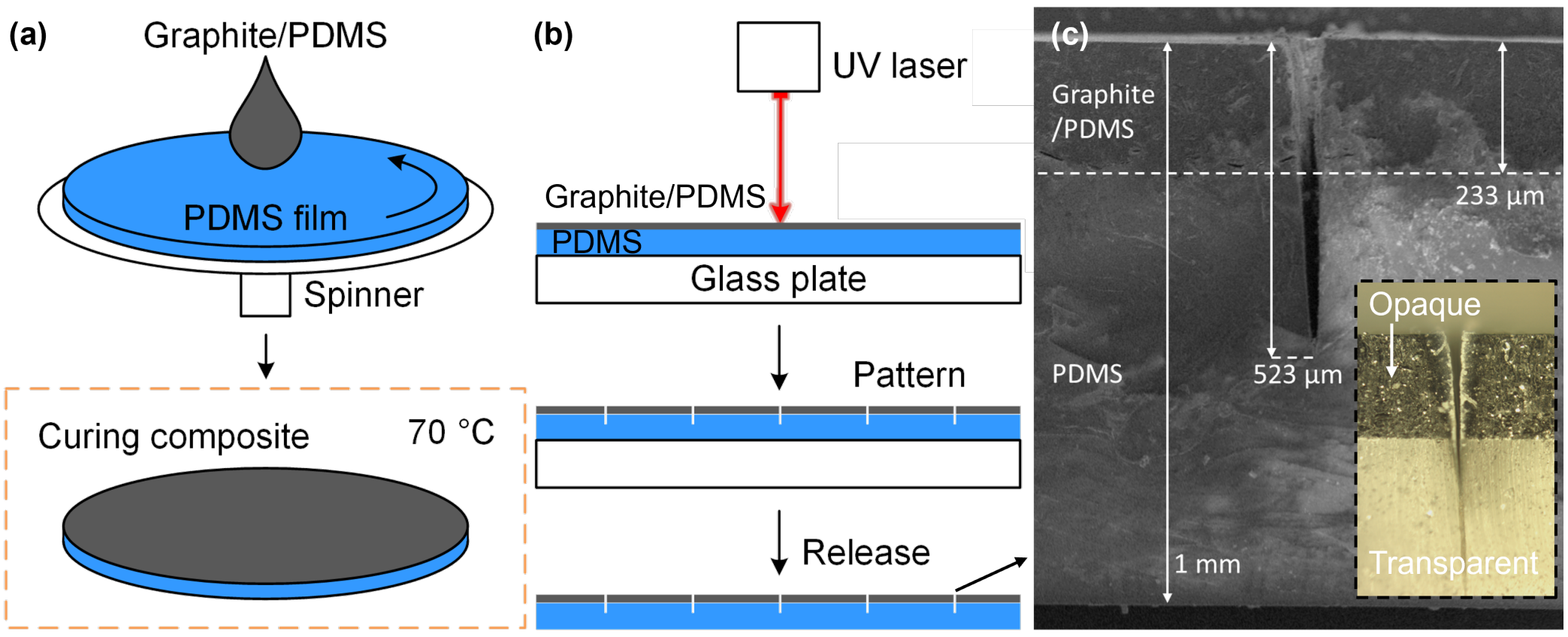

**Figure S4**: Fabrication processes and characterisation. (a) preparation of the film with the spinning coating process. (b) micromachining with high-precision laser cutting. (c) Characterisation of a micro trench with SEM and optical microscope.

**Supplementary Figure S5**

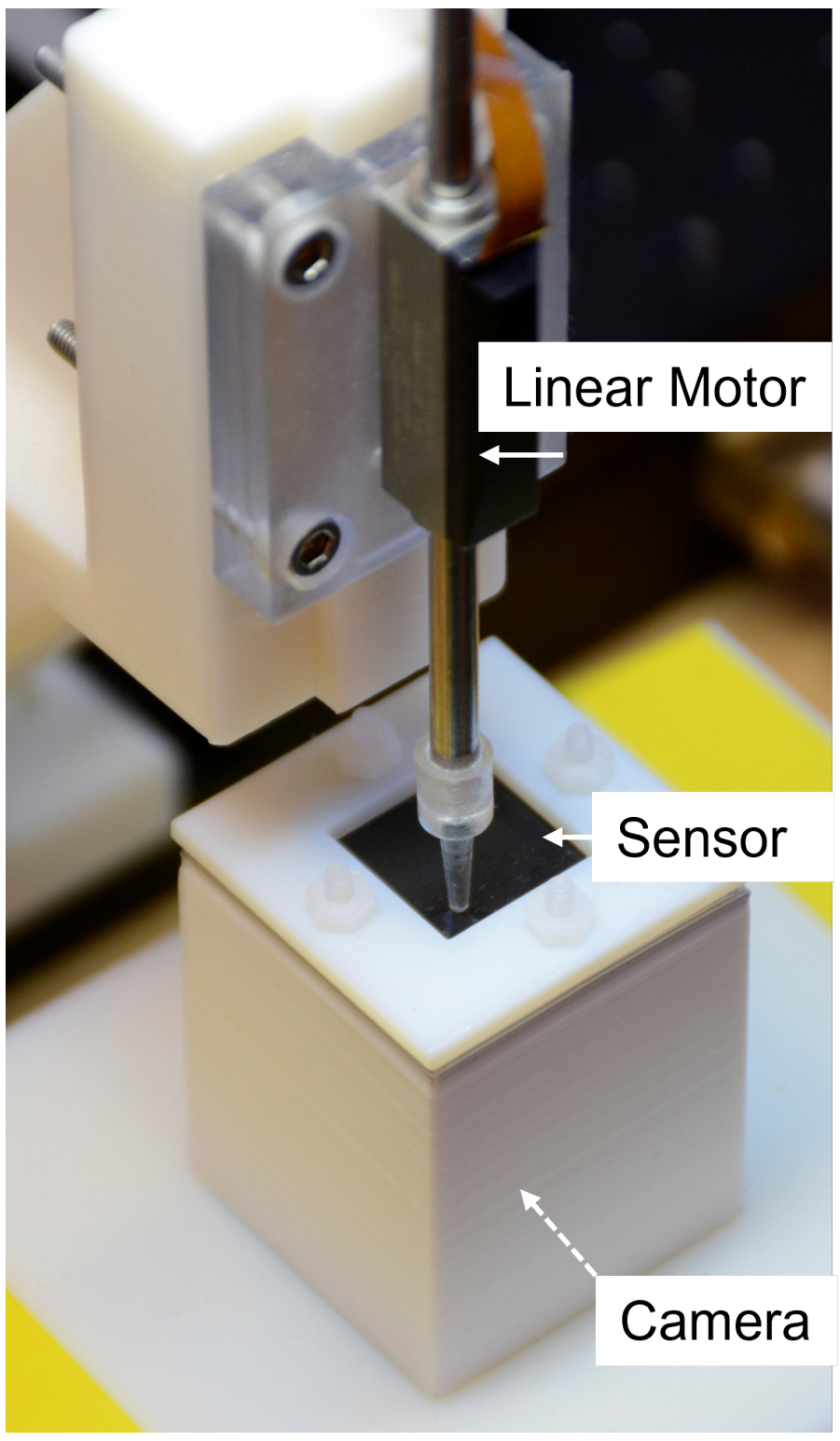

**Figure S5**: Experimental set-up. The setup includes a linear motor (Faulhaber LM083004001), a sensor, and a camera. Diameter of contact area: 1 mm.

**Supplementary**

## Table S1

**Table S1**: Metrics for evaluating model performance

| Model \ Metrics | z, e.g. displacement | | | |
|---|---|---|---|---|
| | MSE | RMSE | MAE | $R^2$ |
| CNN_1 | 0.002225 | 0.047166 | 0.030523 | 0.988868 |
| CNN_3 | 0.002045 | 0.045216 | 0.027137 | 0.989769 |
| CNN_5 | 0.002002 | 0.044741 | 0.026321 | 0.989983 |
| CNN_7 | 0.001955 | 0.04422 | 0.025705 | 0.990215 |
| Model \ Metrics | x-coordinate | | | |
| | MSE | RMSE | MAE | $R^2$ |
| CNN_1 | 0.000553 | 0.023523 | 0.018525 | 0.999723 |
| CNN_3 | 0.00004 | 0.006314 | 0.004997 | 0.99998 |
| CNN_5 | 0.000021 | 0.004588 | 0.003631 | 0.999989 |
| CNN_7 | 0.000007 | 0.002654 | 0.00205 | 0.999996 |
| Model \ Metrics | y-coordinate | | | |
| | MSE | RMSE | MAE | $R^2$ |
| CNN_1 | 0.000873 | 0.029544 | 0.022877 | 0.999564 |
| CNN_3 | 0.000078 | 0.008838 | 0.006963 | 0.999961 |
| CNN_5 | 0.000017 | 0.004092 | 0.003249 | 0.999992 |
| CNN_7 | 0.000006 | 0.002523 | 0.001991 | 0.999997 |

## Supplementary Table S2

**Table S2** Comparison with other works

| Reference | Optical System | Image Resolution | Processing Model | Localization Error | Force/Depth Error |
|---|---|---|---|---|---|
| Duong [12] | 2 fish-eye cameras | 640×480 | Finite element model | Mesh 18×9.5mm | >7mm depth error |
| Gomes [18] | M12 wide-lens, LEDs | / | Projection model | Average 5mm | / |
| Yuan [19] | 1 web camera, LEDs | 640×480 | CNN | / | RMSE<1.9 |
| Fu [26] | 1 wide-angle camera | 1280×720 | Statistical fitting | 6.22-7.19 mm | 9.3%-11.1% |
| This work | 1 web camera | 1280×720 | Lightweight CNN | MAE<0.04 mm, mesh 2×2 mm | MAE<0.03mm |